\documentclass{article}

\makeatletter
\def\input@path{{icml2026/}}
\makeatother
\usepackage{microtype}
\usepackage{graphicx}
\usepackage{booktabs}
\usepackage{array}
\usepackage{tabularx}
\usepackage{ragged2e}
\usepackage{multirow}
\usepackage{amsmath}
\usepackage{amssymb}
\usepackage{mathtools}
\usepackage{hyperref}
\usepackage{subcaption}
\usepackage{placeins}
\PassOptionsToPackage{table}{xcolor}
\usepackage{tikz}
\usetikzlibrary{arrows.meta,positioning,shapes.geometric}

\usepackage[accepted]{icml2026_GenAICreativity}
\definecolor{panelbg}{RGB}{248,248,248}
\definecolor{panelrule}{RGB}{190,190,190}
\definecolor{taskaccent}{RGB}{37,84,124}
\definecolor{failureaccent}{RGB}{150,66,61}
\definecolor{successaccent}{RGB}{48,95,126}
\definecolor{tagbg}{RGB}{239,242,245}
\definecolor{tasklight}{RGB}{232,241,252}
\definecolor{taskgrid}{RGB}{22,112,228}
\definecolor{taskheader}{RGB}{185,216,248}
\definecolor{taskcell}{RGB}{255,255,255}
\definecolor{fileborder}{RGB}{124,151,178}
\definecolor{fileinactive}{RGB}{250,252,255}
\definecolor{failurelight}{RGB}{252,238,236}
\definecolor{successlight}{RGB}{232,244,238}
\definecolor{stderrgray}{RGB}{85,85,85}

\newcommand{\datasetstag}{\makebox[0.92\linewidth][l]{\scriptsize\bfseries Datasets}}
\newcommand{\activefiletag}[1]{\fcolorbox{taskgrid}{taskcell}{\makebox[0.92\linewidth][l]{\hspace{0.25em}\tiny\texttt{#1}}}}
\newcommand{\inactivefiletag}[1]{\fcolorbox{fileborder}{fileinactive}{\makebox[0.92\linewidth][l]{\hspace{0.25em}\tiny\texttt{#1}}}}
\newcommand{\morefiletag}{\fcolorbox{fileborder}{fileinactive}{\makebox[0.38\linewidth][c]{\tiny\ldots}}}

\icmltitlerunning{Searching for Synergy}

\begin{document}

\twocolumn[
\icmltitle{Searching for Synergy in Shared Workspace Human-AI Collaboration}

  \begin{icmlauthorlist}
    \icmlauthor{Nachiket Kotalwar}{cmu}
    \icmlauthor{Rohini Das}{cmu}
    \icmlauthor{Carolyn Ros\'e}{cmu}
  \end{icmlauthorlist}

  \icmlaffiliation{cmu}{Carnegie Mellon University}
  \icmlcorrespondingauthor{Nachiket Kotalwar}{nkotalwa@cs.cmu.edu}

  \icmlkeywords{Human-AI Collaboration, Collaborative Agents, Group Processes, Scientific Discovery}

  \vskip 0.3in
]

\printAffiliationsAndNotice{}

\begin{abstract}
Automated AI agents are increasingly capable, yet many scientific and professional tasks require human judgment and contextual expertise. We use simulated shared-workspace human--AI teams as a controlled testbed for studying how collaboration structure shapes team behavior. Using the Collaborative Gym environment with tasks from DiscoveryBench, we vary team compositions and collaboration structures across 1,482 sessions. We find that adding additional collaborators can lower performance when coordination structure is absent. We then evaluate collaboration scaffolding that combines shared group memory with simulated human-in-the-loop (HITL) gates, where selected actions require approval from a designated simulated participant. This scaffolding improves performance, most clearly in three-person teams, with clearer responsibility signals and stronger routing of expertise to team actions. Overall, our results suggest that coordination structure is central to whether available capability improves team outcomes.
\end{abstract}

\section{Introduction}

Most evaluations of AI agents ask whether an autonomous agent can complete a task. Collaboration asks a different question: can a team turn complementary expertise into a better joint outcome? Live-human studies of long shared-workspace tasks are costly, variable, and hard to control counterfactually, so we use simulated shared-workspace human--AI teams as a controlled testbed for process loss and collaboration scaffolds. This lets us hold the task, model, interface, and profile guidance fixed while varying team composition and collaboration structure. For instance, in data-analysis tasks, a collaborator with
domain expertise may recognize the relevant variable or notice weak evidence, but for that expertise to help, the team
must surface it at the right time, route it to the right decision, and carry it into the final product. If this coordination
breaks down, adding collaborators with relevant expertise can increase interaction without improving outcome.

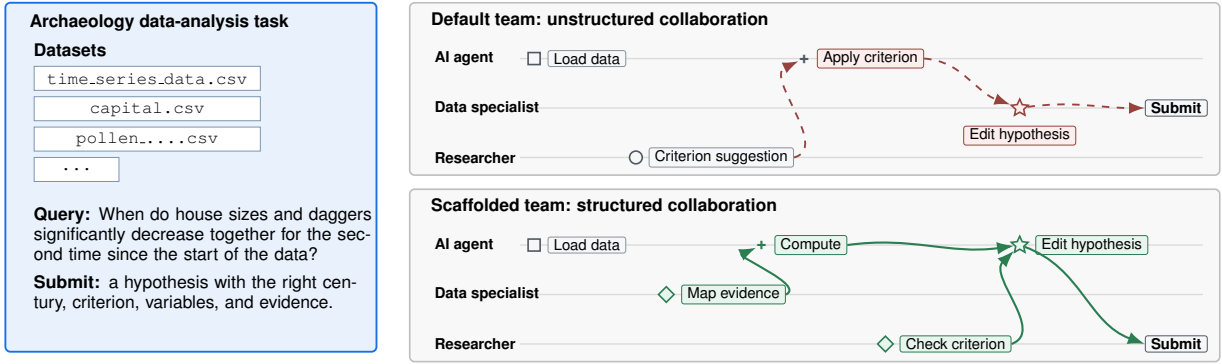
\begin{figure*}[t]
\centering
\resizebox{\textwidth}{!}{%
\definecolor{figonepanelbg}{RGB}{248,248,248}
\definecolor{figonepanelrule}{RGB}{190,190,190}
\definecolor{figonefailure}{RGB}{150,66,61}
\definecolor{figonesuccess}{RGB}{43,125,82}
\definecolor{figoneneutral}{RGB}{80,88,97}
\definecolor{figonetasklight}{RGB}{232,241,252}
\definecolor{figonetaskgrid}{RGB}{22,112,228}
\definecolor{figonetaskcell}{RGB}{255,255,255}
\definecolor{figonefileborder}{RGB}{124,151,178}
\definecolor{figonefailurelight}{RGB}{252,238,236}
\definecolor{figonesuccesslight}{RGB}{232,244,238}
\definecolor{figoneneutrallight}{RGB}{246,247,249}

\begin{tikzpicture}[
  x=1cm,
  y=1cm,
  font=\sffamily\footnotesize,
  outer/.style={draw=figonepanelrule, fill=figonepanelbg, rounded corners=2pt, line width=0.65pt},
  taskouter/.style={draw=figonetaskgrid, fill=figonetasklight, rounded corners=2pt, line width=0.75pt},
  paneltitle/.style={font=\sffamily\scriptsize\bfseries, anchor=west},
  lane/.style={draw=none, font=\sffamily\tiny\bfseries, anchor=west, align=left},
  laneline/.style={draw=figonepanelrule!55, line width=0.55pt},
  filebox/.style={draw=figonefileborder, fill=figonetaskcell, minimum height=0.24cm, inner xsep=2pt, inner ysep=0.6pt, font=\ttfamily\tiny, anchor=west},
  evlabel/.style={draw=figonepanelrule, fill=figonetaskcell, rounded corners=1pt, inner xsep=2.3pt, inner ysep=1pt, font=\sffamily\tiny},
  inspect/.style={draw=figoneneutral, fill=figoneneutrallight, rectangle, minimum size=0.17cm, inner sep=0pt, line width=0.65pt},
  analyze/.style={font=\sffamily\scriptsize\bfseries, text=figoneneutral, inner sep=0pt},
  message/.style={draw=figoneneutral, fill=figoneneutrallight, circle, minimum size=0.18cm, inner sep=0pt, line width=0.65pt},
  check/.style={draw=figoneneutral, fill=figoneneutrallight, diamond, aspect=1.1, minimum size=0.23cm, inner sep=0pt, line width=0.65pt},
  edit/.style={draw=figoneneutral, fill=figoneneutrallight, star, star points=5, star point ratio=2.1, minimum size=0.25cm, inner sep=0pt, line width=0.65pt},
  neutralLabel/.style={evlabel, draw=figoneneutral!65, fill=figoneneutrallight},
  badlabel/.style={evlabel, draw=figonefailure, fill=figonefailurelight},
  goodlabel/.style={evlabel, draw=figonesuccess, fill=figonesuccesslight},
  submitlabel/.style={evlabel, draw=figoneneutral, fill=figoneneutrallight, font=\sffamily\tiny\bfseries},
  badarr/.style={-Latex, draw=figonefailure, dashed, line width=0.75pt},
  goodarr/.style={-Latex, draw=figonesuccess, line width=0.85pt}
]

\draw[taskouter] (0,0.28) rectangle (5.25,5.22);
\node[paneltitle] at (0.22,4.94) {Archaeology data-analysis task};
\node[font=\sffamily\scriptsize\bfseries, anchor=west] at (0.28,4.55) {Datasets};
\node[filebox, minimum width=3.2cm, minimum height=0.34cm, font=\ttfamily\scriptsize] at (0.40,4.15) {time\_series\_data.csv};
\node[filebox, minimum width=3.2cm, minimum height=0.34cm, font=\ttfamily\scriptsize] at (0.40,3.72) {capital.csv};
	\node[filebox, minimum width=3.2cm, minimum height=0.34cm, font=\ttfamily\scriptsize] at (0.40,3.29) {pollen\_\ldots.csv};
\node[filebox, minimum width=1.2cm, minimum height=0.34cm, font=\ttfamily\scriptsize] at (0.40,2.86) {\ldots};

	\node[anchor=west, font=\sffamily\scriptsize, align=left, text width=4.78cm] at (0.28,1.95)
	  {\textbf{Query:} When do house sizes and daggers significantly decrease together for the second time since the start of the data?};
\node[anchor=west, font=\sffamily\scriptsize, align=left, text width=4.78cm] at (0.28,1.10)
  {\textbf{Submit:} a hypothesis with the right century, criterion, variables, and evidence.};

\draw[outer] (5.72,2.78) rectangle (17.18,5.22);
\node[paneltitle] at (5.90,5.00) {Default team: unstructured collaboration};
\foreach \y in {4.43,3.73,3.03} {\draw[laneline] (7.28,\y) -- (16.94,\y);}
\node[lane] at (5.95,4.43) {AI agent};
\node[lane] at (5.95,3.73) {Data specialist};
\node[lane] at (5.95,3.03) {Researcher};

\node[inspect] (bload) at (7.48,4.43) {};
\node[neutralLabel, anchor=west] at (7.67,4.43) {Load data};
\node[message] (bcrit) at (8.92,3.03) {};
\node[neutralLabel, anchor=west] (bcritlab) at (9.10,3.03) {Criterion suggestion};
\node[analyze] (bwrong) at (11.30,4.43) {+};
\node[badlabel, anchor=west] (bwronglab) at (11.48,4.43) {Apply criterion};
\node[edit, draw=figonefailure, fill=figonefailurelight] (bedit) at (14.35,3.73) {};
\node[badlabel, anchor=north] at (14.35,3.48) {Edit hypothesis};
\node[submitlabel, anchor=west] (bsub) at (16.12,3.73) {Submit};
\draw[badarr] (bcritlab.east) to[out=15,in=205] (bwrong);
\draw[badarr] (bwronglab.east) to[out=-4,in=160] (bedit);
\draw[badarr] (bedit) to[out=8,in=180] (bsub.west);

\draw[outer] (5.72,0.14) rectangle (17.18,2.58);
\node[paneltitle] at (5.90,2.36) {Scaffolded team: structured collaboration};
\foreach \y in {1.79,1.09,0.39} {\draw[laneline] (7.28,\y) -- (16.94,\y);}
\node[lane] at (5.95,1.79) {AI agent};
\node[lane] at (5.95,1.09) {Data specialist};
\node[lane] at (5.95,0.39) {Researcher};

\node[inspect] (sload) at (7.48,1.79) {};
\node[neutralLabel, anchor=west] at (7.67,1.79) {Load data};
	\node[check, draw=figonesuccess, fill=figonesuccesslight] (sdata) at (9.35,1.09) {};
	\node[goodlabel, anchor=west] (sdatalab) at (9.56,1.09) {Map evidence};
\node[analyze, text=figonesuccess] (scomp) at (10.70,1.79) {+};
\node[goodlabel, anchor=west] (scomplab) at (10.88,1.79) {Compute};
\node[check, draw=figonesuccess, fill=figonesuccesslight] (sresearch) at (12.45,0.39) {};
\node[goodlabel, anchor=west] (sresearchlab) at (12.65,0.39) {Check criterion};
\node[edit, draw=figonesuccess, fill=figonesuccesslight] (sedit) at (14.35,1.79) {};
\node[goodlabel, anchor=west] at (14.57,1.79) {Edit hypothesis};
	\node[submitlabel, anchor=west] (ssub) at (16.12,0.39) {Submit};
	\draw[goodarr] (sdatalab.east) to[out=25,in=205] (scomp);
	\draw[goodarr] (sresearchlab.east) to[out=20,in=225] (sedit);
	\draw[goodarr] (scomplab.east) to[out=8,in=188] (sedit);
	\draw[goodarr] (sedit) to[out=-20,in=160] (ssub.west);

\end{tikzpicture}}
\caption{How collaboration structure changes the path to submission. The task requires a submitted hypothesis with the relevant variables, temporal criterion, and supporting evidence. The Data specialist and Researcher lanes denote simulated-human collaborator profiles. In the Default trace, collaborator input is not converted into a check on the agent's edit, so a criterion suggestion is misapplied and carried into an unsupported hypothesis edit. In the Scaffolded trace, shared group memory and simulated HITL gates route the relevant checks before computation and editing. The panels summarize events from the full traces.}
\label{fig:task-collaboration-overview}
\end{figure*}

Classic group-process research calls this process loss: teams fail to convert member resources into productivity when coordination is ineffective \citep{steiner1972group}. Coordination theory frames collaboration as managing dependencies among activities \citep{malone1994coordination}, and coordination neglect describes how teams underweight the integration work that interdependent contributions require \citep{heath2000coordination}. Human--AI teams show analogous patterns: adding AI or human expertise is not always beneficial on average \citep{vaccaro2024humanai}, people may over-rely on or misread AI recommendations despite explanations \citep{bansal2021whole}, and professional AI assistance can have limited average benefit when humans and AI do not combine their signals effectively \citep{agarwal2023radiology,yu2024heterogeneity}.

We therefore use group-process research as a design lens for simulated shared-workspace human--AI teams. We evaluate two collaboration structures: shared group memory, which externalizes who knows what, what evidence has been established, and what checks remain; and simulated HITL gates, which require selected actions to be approved by a designated participant. These structures test whether making expertise, responsibility, and evidence checks explicit can reduce process loss in collaborative agent trajectories. Because such failures may appear in the interaction before they appear in the final answer, we evaluate both submitted hypotheses and process traces.

We focus on shared-workspace collaboration, where participants share tools, messages, and artifacts and decide during the task who does what \citep{shao2024cogym}. Within this simulation framework, we study collaborative data analysis using archaeology tasks from DiscoveryBench \citep{majumder2024discoverybench}, a multi-step discovery benchmark where tasks require domain interpretation, variable mapping, and evidence construction (Figure~\ref{fig:task-collaboration-overview}).

The key empirical pattern is counterintuitive: adding additional relevant collaborators can reduce performance when teams lack the structure to coordinate contributions. Trace diagnostics point to unassigned responsibilities and weak evidence handoff rather than missing capability alone. We then evaluate a scaffolded setting that combines shared group memory with simulated HITL gates; with this scaffolding in place, teams show more distributed initiative, clearer responsibility signals, and higher mean performance, most clearly in the three-person setting.

Our contributions are:
\begin{enumerate}
    \item We extend Collaborative Gym for human--AI team studies on DiscoveryBench, enabling systematic variation in team composition and collaboration structure.
    \item We document a process-loss pattern in simulated shared-workspace teams and use trace-level diagnostics to examine how useful intermediate work fails to reach the submitted hypothesis.
    \item We evaluate scaffolded collaboration with shared group memory and simulated HITL gates, showing higher mean performance in simulated teams, most clearly in the three-person setting, alongside trace-level evidence of responsibility assignment, routed checks, and stronger evidence grounding.
\end{enumerate}

\section{Related Work}

\paragraph{Collaborative-agent benchmarks.}
Recent interactive-agent environments study collaboration beyond isolated answer generation. Some emphasize social simulation and believable role-played interaction \citep{park2023generativeagents,zhou2024sotopia}; others test proactive assistance through active user simulation \citep{nathani2026pare}, or workflow orchestration under coupling, asynchrony, temporal constraints, and time-efficiency objectives \citep{masters2025manageragentgym,sun2025collabovercooked,gonzalezpumariega2025robotouille,lin2024asynchow,paracook2025}. These largely study turn-based or task-orchestration settings rather than open shared-workspace coordination. Collaborative Gym provides a shared-workspace testbed for human--AI collaboration with communication, tool use, artifact editing, and flexible non-turn-taking interaction \citep{shao2024cogym}, which connects to workspace-awareness research on how collaborators stay informed of one another's actions in joint work \citep{gutwin2002workspace}. We build on Collaborative Gym and DiscoveryBench \citep{majumder2024discoverybench} by varying team size and giving simulated collaborators distinct private guidance, then asking whether teams turn distributed evidence into a supported scientific hypothesis.

\paragraph{Group memory and shared understanding.}
Our group-memory scaffold targets process loss by making distributed knowledge easier to locate and use \citep{steiner1972group}. This motivation also matches classic findings on biased information sampling: groups tend to discuss shared information more readily than unshared information, limiting the value of distributed expertise \citep{stasser1985pooling}. It draws on transactive memory systems, which describe how teams coordinate distributed knowledge through shared awareness of who knows what and how that knowledge should be retrieved during work \citep{wegner1987transactive,moreland1999transactive,lewis2003tms,lewis2011tms,argote2012tms}. Shared mental models similarly connect team effectiveness to common expectations about the task, teammates, and workflow, including in human--AI teams \citep{mathieu2000sharedmentalmodels,andrews2022sharedmentalmodels}.

\paragraph{Simulated HITL gates and collaborative control.}
Simulated HITL gates draw on collaboration scripts, which use external structure to guide participation in joint work through responsibilities, activity sequences, and interaction moves \citep{kollar2006collaboration,fischer2013script}. They also relate to mixed-initiative and automation-control work on when automated systems should act, defer, or solicit human input \citep{horvitz1999mixedinitiative,parasuraman2000automation,amershi2019guidelines}. Recent agent evaluations use HITL-style abstractions without live human participants in the evaluated loop, including targeted gates in autonomous research systems and an \texttt{ask\_human()} tool for selective escalation \citep{liu2026autoresearchclaw,elfeki2026hilbench}. In the AutoResearchClaw HITL ablation, targeted intervention outperformed dense step-by-step oversight \citep{liu2026autoresearchclaw}, consistent with the idea that the routing of human input matters, not only its frequency. We adapt this gating pattern so that the team itself decides which actions need sign-off and from whom.

\section{Problem Setup}

We follow the tabular-analysis setup from Collaborative Gym \citep{shao2024cogym}. Each task instance is a tuple
\[
x = (\mathcal{D}, q, y^\star),
\]
where $\mathcal{D}$ is a set of CSV files, $q$ is a natural-language query, and $y^\star$ is the benchmark reference hypothesis. A team inspects the data, communicates in a shared workspace, can run analysis code, and submits a hypothesis through the result editor. We denote this submitted hypothesis by $\hat{y}$; it should identify the relevant task context, map the query to the correct variables, state the target relationship, and justify the claim with evidence from the data.

A session has participants $\mathcal{U}$ and produces an ordered interaction trace
\[
\tau = ((u_t, a_t, o_t))_{t=1}^{T},
\]
where $u_t \in \mathcal{U}$ is the participant acting at event $t$, $a_t$ is the action, and $o_t$ is the resulting observation. Each participant $u$ has fixed private guidance $\pi_u$ that defines its collaborator profile, such as data-analysis or researcher guidance. When the acting participant is $u=u_t$, its action is
\[
a_t = f_u\!\left(x, o_{t-1}, M_{<t}, A^u_{<t}, \pi_u\right),
\]
where $M_{<t}$ is the chat history and $A^u_{<t}$ is the action history visible to participant $u$. In our implementation, the AI agent sees the team-level action history, while simulated human collaborators see only their own action history rather than the full team history \citep{nathani2026pare}. We therefore evaluate both the hypothesis $\hat{y}$ and the process trace $\tau$, since useful intermediate work in $\tau$ may never reach $\hat{y}$ (Section~\ref{sec:metrics}).

\section{Experimental Setup}
\label{sec:experimental-setup}

\subsection{Overview}
\label{sec:study-design}
We vary team composition and collaboration structure to test whether teams turn available expertise into well-supported hypotheses. Sections~\ref{sec:task-suite}--\ref{sec:team-compositions-combinations} define the tasks, participants, structures, and team-composition variants. We evaluate submissions with the task Performance score and trace-level activity/process metrics (Section~\ref{sec:metrics}); Section~\ref{sec:results} reports aggregate results.

\subsection{Task Suite}
\label{sec:task-suite}
We use the complete 38-task archaeology subset of DiscoveryBench \citep{majumder2024discoverybench}. This subset is well suited to collaborative data analysis because its tasks require domain-specific interpretation, careful variable mapping, and multi-step evidence construction. Collaborators can help by clarifying archaeological terms, selecting variables, reviewing temporal criteria, or verifying support for a proposed hypothesis. We use one complete domain subset rather than pooling all DiscoveryBench domains so that task semantics, data conventions, and evaluator expectations remain comparable across team compositions and collaboration structures.

\subsection{Participants}
\label{sec:participants-model-control}
Each session includes one AI agent and zero to two simulated human collaborators, depending on team composition. All participants use DeepSeek V3.2 with Collaborative Gym's ReAct-style action loop and private scratchpad memory \citep{yao2023react,shao2024cogym}; \citet{shao2024cogym} report that these simulated collaborators reproduce key behavioral patterns of real participants. We use these simulators as controlled collaborator profiles for comparing collaboration structures at scale. All participants share one model and interface, and only the profile guidance $\pi_u$ differs. Differences across variants therefore reflect team composition and collaboration structure, not model or interface.

\subsection{Collaboration Structures}
\label{sec:collaboration-structures}
Our main comparison is between Default and Scaffolded collaboration. Default is the original Collaborative Gym shared-workspace setup without additional coordination mechanisms. Scaffolded collaboration combines shared group memory with simulated HITL gates: the team first builds a shared record of expertise, responsibilities, work plan, and evidence criteria, then uses that record to decide which actions require approval and who owns those approvals. We also evaluate two diagnostic variants: shared group memory only, and preassigned simulated HITL gates whose owners are configured by action type rather than chosen by the team.

These structures scaffold coordination without changing the task suite, underlying LLM, action interface, profile guidance, or domain knowledge. Simulated HITL gates add explicit approval responsibility for selected actions, while shared group memory adds a shared coordination artifact. The scaffolds deliberately change how much context participants share and how often they can communicate.

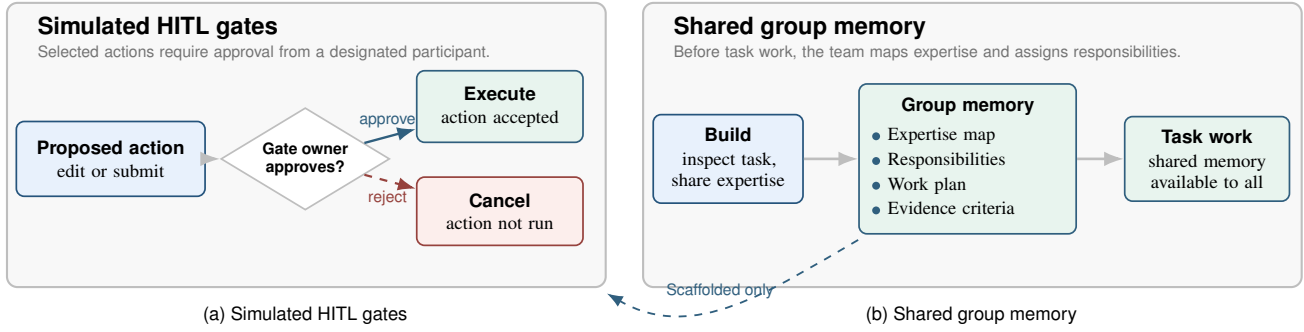
\begin{figure*}[t]
\centering
\resizebox{\textwidth}{!}{%
\begin{tikzpicture}[
  font=\sffamily\scriptsize,
  panel/.style={draw=panelrule, fill=panelbg, rounded corners=4pt, line width=0.8pt},
  box/.style={draw=panelrule, fill=white, rounded corners=3pt, line width=0.65pt, inner sep=6pt, align=center},
  bluebox/.style={box, draw=taskaccent, fill=tasklight},
  greenbox/.style={box, draw=successaccent, fill=successlight},
  redbox/.style={box, draw=failureaccent, fill=failurelight},
  arrow/.style={-Latex, draw=panelrule, line width=1.0pt},
  goodarrow/.style={-Latex, draw=successaccent, line width=0.9pt},
  badarrow/.style={-Latex, draw=failureaccent, line width=0.85pt, dashed}
]

\node[panel, minimum width=8.0cm, minimum height=3.8cm, anchor=north west] (pa) at (0,3.8) {};
\node[anchor=west, font=\sffamily\footnotesize\bfseries] at (0.30,3.45) {Simulated HITL gates};
\node[anchor=west, text=black!55, font=\sffamily\tiny, text width=7.2cm] at (0.30,3.12) {Selected actions require approval from a designated participant.};

  \node[bluebox, text width=2.1cm, minimum height=1.0cm] (proposal) at (1.40,1.70) {\bfseries Proposed action\\[0.15em]\normalfont edit or submit};
\node[draw=panelrule, fill=white, diamond, aspect=1.65, inner sep=2pt, line width=0.65pt, font=\sffamily\tiny\bfseries, align=center] (decision) at (4.00,1.70) {Gate owner\\approves?};
\node[greenbox, text width=1.8cm, minimum height=0.8cm] (execute) at (6.60,2.40) {\bfseries Execute\\[0.1em]\normalfont action accepted};
\node[redbox, text width=1.8cm, minimum height=0.8cm] (cancel) at (6.60,1.00) {\bfseries Cancel\\[0.1em]\normalfont action not run};

\draw[arrow] (proposal) -- (decision);
\draw[goodarrow] (decision) -- node[above, font=\sffamily\tiny, text=successaccent, pos=0.45] {approve} (execute);
\draw[badarrow] (decision) -- node[below, font=\sffamily\tiny, text=failureaccent, pos=0.45] {reject} (cancel);

\node[font=\sffamily\scriptsize, anchor=north] at (4.0,-0.15) {(a) Simulated HITL gates};

\node[panel, minimum width=8.8cm, minimum height=3.8cm, anchor=north west] (pb) at (8.5,3.8) {};
\node[anchor=west, font=\sffamily\footnotesize\bfseries] at (8.80,3.45) {Shared group memory};
\node[anchor=west, text=black!55, font=\sffamily\tiny, text width=8.0cm] at (8.80,3.12) {Before task work, the team maps expertise and assigns responsibilities.};

\node[bluebox, text width=1.65cm, minimum height=1.0cm, inner sep=5pt] (discover) at (9.65,1.70) {\bfseries Build\\[0.15em]\normalfont inspect task,\\share expertise};

\node[greenbox, text width=2.55cm, inner sep=5pt, align=left] (memory) at (12.85,1.70) {
  \centering\bfseries Group memory\\[0.45em]
  \normalfont\raggedright
  \begin{tabular}{@{}c@{\hskip 3pt}l@{}}
  \tikz\fill[successaccent] (0,0) circle (1.5pt); & Expertise map\\[0.2em]
  \tikz\fill[successaccent] (0,0) circle (1.5pt); & Responsibilities\\[0.2em]
  \tikz\fill[successaccent] (0,0) circle (1.5pt); & Work plan\\[0.2em]
  \tikz\fill[successaccent] (0,0) circle (1.5pt); & Evidence criteria\\
  \end{tabular}
};

\node[greenbox, text width=1.85cm, minimum height=1.0cm, inner sep=5pt] (context) at (16.05,1.70) {\bfseries Task work\\[0.15em]\normalfont shared memory\\available to all};

\draw[arrow] (discover.east) -- (memory.west);
\draw[arrow] (memory.east) -- (context.west);

\draw[draw=successaccent, line width=0.7pt, dashed, -Latex] ([yshift=-2pt]memory.south west) to[out=210,in=-30] node[above, font=\sffamily\tiny, text=successaccent, pos=0.55] {Scaffolded only} ([yshift=-2pt]pa.south east);

\node[font=\sffamily\scriptsize, anchor=north] at (12.90,-0.15) {(b) Shared group memory};

\end{tikzpicture}}
\caption{Collaboration structures used in the study. (a) Simulated HITL gates require approval from a designated participant before selected actions take effect. (b) Shared group memory is built through pre-task discussion; the team maps expertise, responsibilities, evidence criteria, and a work plan. In the Scaffolded setting, this shared memory helps determine which actions are gated and who owns each gate in (a); in the diagnostic preassigned-gates variant, owners are configured by action type.}
\label{fig:collaboration-scaffolds}
\end{figure*}

\paragraph{Simulated HITL gates.}
Simulated HITL gates designate selected actions as requiring approval from a specific participant before they take effect (Figure~\ref{fig:collaboration-scaffolds}a). The approving participant is simulated in our experimental setup. Not all actions require approval: in the Scaffolded setting, the team decides which actions need gates and who should approve them, based on the expertise and responsibilities they have mapped; actions without a designated gate owner proceed normally. This abstracts common approval patterns in collaborative work, where code review, clinical sign-off, and AI coding assistants route consequential operations through a designated approver while letting routine work flow. As a diagnostic variant, we also evaluate preassigned simulated HITL gates where owners are configured by action type rather than team-chosen.

\paragraph{Shared group memory.}
Shared group memory adds a pre-task build phase based on transactive memory systems in group-process research \citep{wegner1987transactive,moreland1999transactive,lewis2011tms}. The team records who knows what, who should be trusted for what, how work should be coordinated, and what evidence criteria the final answer must satisfy (Figure~\ref{fig:collaboration-scaffolds}b). Participants inspect the task context, discuss and revise entries, and agree on this map. Unlike each participant's private memory (Section~\ref{sec:participants-model-control}), it is shared team state: a single expertise-and-responsibility map available to every participant. Once the build phase ends, the memory becomes a fixed reference that participants consult but do not update.

\paragraph{Scaffolded collaboration.}
The two scaffolds work together: the group-memory build phase is where the team decides which actions need approval and who owns each one (leaving some actions ungated), and the gates then enforce those decisions during the task. When a designated action is proposed, the chosen owner must approve or reject it before it takes effect, and the agreed expertise map and criteria provide context for that decision.

\subsection{Team Compositions}
\label{sec:team-compositions-combinations}
Table~\ref{tab:team-compositions-structures} lists the evaluated team variants. Suffixes denote simulated-human profile guidance, not guaranteed competence: \texttt{D} is a data-analysis collaborator profile focused on variable mapping, filters, computation, and numeric evidence; \texttt{R} is a researcher collaborator profile focused on query semantics, relation wording, ambiguity, caveats, and evidential support; and \texttt{DR} denotes a team with both simulated-human collaborators, one with each profile. Labels join structure and suffix, such as Default-D or Scaffolded-DR. Appendix~\ref{app:simulated-human-guidance} gives the fuller profile descriptions.

\begin{table*}[t]
\centering
\small
\caption{Team-composition and collaboration-structure design. Single-agent has no collaborator profile; each collaborative structure is evaluated with D, R, and DR profile variants. Diagnostic variants isolate parts of the Scaffolded setting but are not symmetric ablations because preassigned gates use externally configured owners.}
\label{tab:team-compositions-structures}
\begin{tabularx}{\textwidth}{ll l X}
\toprule
Structure & Profile variants & Team composition & Structure details \\
\midrule
Single-agent & -- & AI only & No simulated-human collaborator. \\
Default & D, R, DR & AI + simulated human(s) & Original shared workspace. \\
Preassigned gates (diagnostic) & D, R, DR & AI + simulated human(s) & Simulated HITL gates; owners by action type. \\
Shared group memory only (diagnostic) & D, R, DR & AI + simulated human(s) & Shared group memory; no approval gates. \\
Scaffolded & D, R, DR & AI + simulated human(s) & Shared group memory and HITL gates; team chooses owners. \\
\bottomrule
\end{tabularx}
\vspace{0.25em}
\footnotesize\emph{Variant legend:} D = data-analysis profile; R = researcher profile; DR = two collaborators, one with each profile.
\end{table*}

\subsection{Evaluation Metrics}
\label{sec:metrics}

The metrics separate properties of the submitted hypothesis $\hat{y}$ from properties of the interaction trace $\tau$. The outcome metric is Collaborative Gym's normalized Task Performance score, reported as \emph{Performance}, where higher is better \citep{shao2024cogym}; evaluator model details are in Appendix~\ref{app:evaluator-models}. Activity metrics count trace actions: Human Work ($W_{\mathrm{human}}$) is the number of non-message actions by simulated humans; Total Work ($W_{\mathrm{total}}$) is the number of non-message actions by all participants; and Team Messages ($M_{\mathrm{team}}$) is the number of message actions by all participants.

For initiative structure, we use Collaborative Gym's Initiative Entropy \citep{shao2024cogym}. We report the normalized form, $H_{\mathrm{init,norm}}$, so scores are comparable across team sizes:
\[
H_{\mathrm{init,norm}} =
\frac{-\sum_{u \in \mathcal{U}} p_u \log p_u}{\log |\mathcal{U}|},
\]
where $\mathcal{U}$ is the team and $p_u$ is the fraction of initiative events attributed to participant $u$. Higher values indicate more evenly distributed initiative.

\paragraph{Reference workflow graphs.}
The graph-dependent process metrics use a reference workflow graph for each task,
\[
G_i=(V_i,E_i,\mathcal{P}_i,\mathcal{C}_i),
\]
where $V_i$ contains reference reasoning and evidence nodes, $E_i$ their dependencies, $\mathcal{P}_i$ acceptable solution paths, and $\mathcal{C}_i$ completion criteria for a supported submission. We construct each graph from the task data, query, and benchmark reference hypothesis, adding alternative evidence routes where applicable. Because $\mathcal{P}_i$ can contain multiple acceptable solution paths, the graphs represent a family of valid evidence-and-reasoning routes rather than a single canonical workflow. Before manual validation, we check each graph for schema validity, executable evidence references, consistency with the reference hypothesis, and well-formed dependencies. The reference workflow graphs and validation annotations are released with our code; the repository link is in Appendix~\ref{app:simulated-human-guidance}.

An LLM judge (model details in Appendix~\ref{app:evaluator-models}) labels trace-level events given the session trace, task information, and validated graph. Workflow Coverage ($C_{\mathrm{wf}}$) measures how much of an acceptable reference path appears in the trace, with nodes labeled satisfied, partially satisfied, or missing and scored as 1, 0.5, or 0 under $\mathcal{P}_i$. Hypothesis Support ($S_{\mathrm{hyp}}$) measures whether $\hat{y}$'s relation, scope, and evidential grounding are supported by trace evidence under $\mathcal{C}_i$. Profile Alignment ($A_{\mathrm{profile}}$), computed only when profiled collaborators are present, measures whether their contributions match the profile and the workflow's needs, not how much they participated.

\section{Results}
\label{sec:results}

\subsection{Overview}
\label{sec:aggregation}
We ask three questions in this simulated setting. First, how do Default collaborative teams compare with the single-agent baseline, and where does process loss appear? Second, do collaboration structures drawn from group-process research recover performance by shifting coordination patterns? Third, what do these structures change at the process level (initiative structure, evidence handoff, or both)?

For each team variant and task, we average metrics across three independent seeds; tables report means with standard errors across tasks. Table~\ref{tab:results-main} reports the headline aggregate results: the single-agent baseline, Default shared-workspace teams, and Scaffolded teams. Table~\ref{tab:appendix-diagnostic-results} reports diagnostic variants that isolate shared group memory and preassigned simulated HITL gates.

\begin{table*}[t]
\caption{Headline aggregate metrics by team composition and collaboration structure. Cells report task-level mean (SE) after averaging across seeds. Activity counts are raw; process and Performance metrics are on $[0,1]$. Dashes mark non-applicable metrics; diagnostic variants appear in Table~\ref{tab:appendix-diagnostic-results}.}
\label{tab:results-main}
\centering
{\setlength\tabcolsep{3.2pt}
\renewcommand{\arraystretch}{1.02}
\resizebox{\textwidth}{!}{%
\begin{tabular}{lrrrrrrrr}
\toprule
\multicolumn{1}{c}{\textbf{Structure / profile}} &
\multicolumn{3}{c}{\textbf{Activity}} &
\multicolumn{4}{c}{\textbf{Process}} &
\multicolumn{1}{c}{\textbf{Outcome}} \\
\cmidrule(r){1-1}
\cmidrule(lr){2-4}
\cmidrule(lr){5-8}
\cmidrule(l){9-9}
\multicolumn{1}{c}{} &
\multicolumn{1}{r}{\small $W_{\mathrm{human}}$} &
\multicolumn{1}{r}{\small $W_{\mathrm{total}}$} &
\multicolumn{1}{r}{\small $M_{\mathrm{team}}$} &
\multicolumn{1}{r}{\small $H_{\mathrm{init,norm}}$} &
\multicolumn{1}{r}{\small $A_{\mathrm{profile}}$} &
\multicolumn{1}{r}{\small $C_{\mathrm{wf}}$} &
\multicolumn{1}{r}{\small $S_{\mathrm{hyp}}$} &
\multicolumn{1}{r}{\small Perf.} \\
\midrule
\multicolumn{9}{l}{\textit{Baseline}} \\
\quad Single-agent &
$-$ &
$5.6$ {\color{stderrgray} $(0.2)$} &
$-$ &
$-$ &
$-$ &
$0.63$ {\color{stderrgray} $(0.04)$} &
$0.28$ {\color{stderrgray} $(0.06)$} &
$0.71$ {\color{stderrgray} $(0.05)$} \\
\midrule
\multicolumn{9}{l}{\textit{Default shared workspace}} \\
\quad D profile &
$0.95$ {\color{stderrgray} $(0.1)$} &
$7.2$ {\color{stderrgray} $(0.3)$} &
$3.5$ {\color{stderrgray} $(0.3)$} &
$0.31$ {\color{stderrgray} $(0.05)$} &
$0.68$ {\color{stderrgray} $(0.04)$} &
$0.65$ {\color{stderrgray} $(0.04)$} &
$0.18$ {\color{stderrgray} $(0.05)$} &
$0.69$ {\color{stderrgray} $(0.06)$} \\
\quad R profile &
$0.62$ {\color{stderrgray} $(0.09)$} &
$7.5$ {\color{stderrgray} $(0.4)$} &
$3.5$ {\color{stderrgray} $(0.3)$} &
$0.37$ {\color{stderrgray} $(0.04)$} &
$0.65$ {\color{stderrgray} $(0.04)$} &
$0.61$ {\color{stderrgray} $(0.04)$} &
$0.18$ {\color{stderrgray} $(0.04)$} &
$0.68$ {\color{stderrgray} $(0.07)$} \\
\quad DR profiles &
$1.6$ {\color{stderrgray} $(0.2)$} &
$7.6$ {\color{stderrgray} $(0.4)$} &
$6.1$ {\color{stderrgray} $(0.5)$} &
$0.54$ {\color{stderrgray} $(0.03)$} &
$0.66$ {\color{stderrgray} $(0.04)$} &
$0.62$ {\color{stderrgray} $(0.04)$} &
$0.19$ {\color{stderrgray} $(0.06)$} &
$0.63$ {\color{stderrgray} $(0.06)$} \\
\midrule
\multicolumn{9}{l}{\textit{Scaffolded}} \\
\quad D profile &
$1.2$ {\color{stderrgray} $(0.1)$} &
$6.9$ {\color{stderrgray} $(0.3)$} &
$11$ {\color{stderrgray} $(0.6)$} &
$0.74$ {\color{stderrgray} $(0.04)$} &
$0.69$ {\color{stderrgray} $(0.04)$} &
$0.63$ {\color{stderrgray} $(0.04)$} &
$0.23$ {\color{stderrgray} $(0.06)$} &
$0.72$ {\color{stderrgray} $(0.05)$} \\
\quad R profile &
$1.4$ {\color{stderrgray} $(0.3)$} &
$7.5$ {\color{stderrgray} $(0.6)$} &
$12$ {\color{stderrgray} $(0.4)$} &
$0.77$ {\color{stderrgray} $(0.03)$} &
$0.66$ {\color{stderrgray} $(0.04)$} &
$0.61$ {\color{stderrgray} $(0.04)$} &
$0.18$ {\color{stderrgray} $(0.05)$} &
$0.73$ {\color{stderrgray} $(0.06)$} \\
\quad DR profiles &
$2.2$ {\color{stderrgray} $(0.2)$} &
$7.9$ {\color{stderrgray} $(0.4)$} &
$18$ {\color{stderrgray} $(0.7)$} &
$0.85$ {\color{stderrgray} $(0.02)$} &
$0.70$ {\color{stderrgray} $(0.04)$} &
$0.65$ {\color{stderrgray} $(0.04)$} &
$0.23$ {\color{stderrgray} $(0.06)$} &
$0.76$ {\color{stderrgray} $(0.05)$} \\
\bottomrule
\end{tabular}}}
\end{table*}

\newcommand{\mainTaskTwentySeven}{%
\fcolorbox{taskgrid}{tasklight}{%
\begin{minipage}[t]{0.94\linewidth}
\setlength{\fboxsep}{1.5pt}
\footnotesize
\begin{minipage}[t]{0.32\linewidth}
\vspace{0pt}
\datasetstag\\[0.18em]
\activefiletag{time\_series\_data.csv}\\[0.22em]
\inactivefiletag{capital.csv}\\[0.22em]
\inactivefiletag{pollen\_\ldots.csv}\\[0.22em]
\morefiletag
\end{minipage}\hfill
\begin{minipage}[t]{0.62\linewidth}
\vspace{0pt}
\scriptsize
\setlength{\tabcolsep}{3.5pt}
\setlength{\arrayrulewidth}{0.45pt}
\renewcommand{\arraystretch}{1.16}
\arrayrulecolor{taskgrid}
\begin{tabularx}{\linewidth}{|>{\centering\arraybackslash}p{0.17\linewidth}|>{\centering\arraybackslash}p{0.24\linewidth}|>{\centering\arraybackslash}p{0.20\linewidth}|>{\centering\arraybackslash}X|}
\hline
\rowcolor{taskheader}\textbf{CE} & \textbf{HouseSize} & \textbf{Dagger} & \textbf{\ldots} \\
\hline
\rowcolor{taskcell}
-4100 & -1.35 & -0.58 & \ldots \\
\hline
\rowcolor{taskcell}
-4000 & -1.35 & -0.58 & \ldots \\
\hline
\rowcolor{taskcell}
-3900 & -1.35 & -0.58 & \ldots \\
\hline
\rowcolor{taskcell}
\ldots & \ldots & \ldots & \ldots \\
\hline
\end{tabularx}
\arrayrulecolor{black}
\setlength{\arrayrulewidth}{0.4pt}
\end{minipage}
\par\vspace{0.45em}\noindent
{\bfseries Goal:} In which century did house sizes and daggers significantly decrease simultaneously for the second time since the start of the observational data?
\end{minipage}}}

\begin{figure*}[!t]
\centering
\setlength{\fboxsep}{4pt}
\begin{subfigure}[t]{0.56\textwidth}
\centering
\mainTaskTwentySeven
\caption{Task specification}
\end{subfigure}

\vspace{0.35em}
\begin{subfigure}[t]{0.88\textwidth}
\centering
\includegraphics[width=\linewidth]{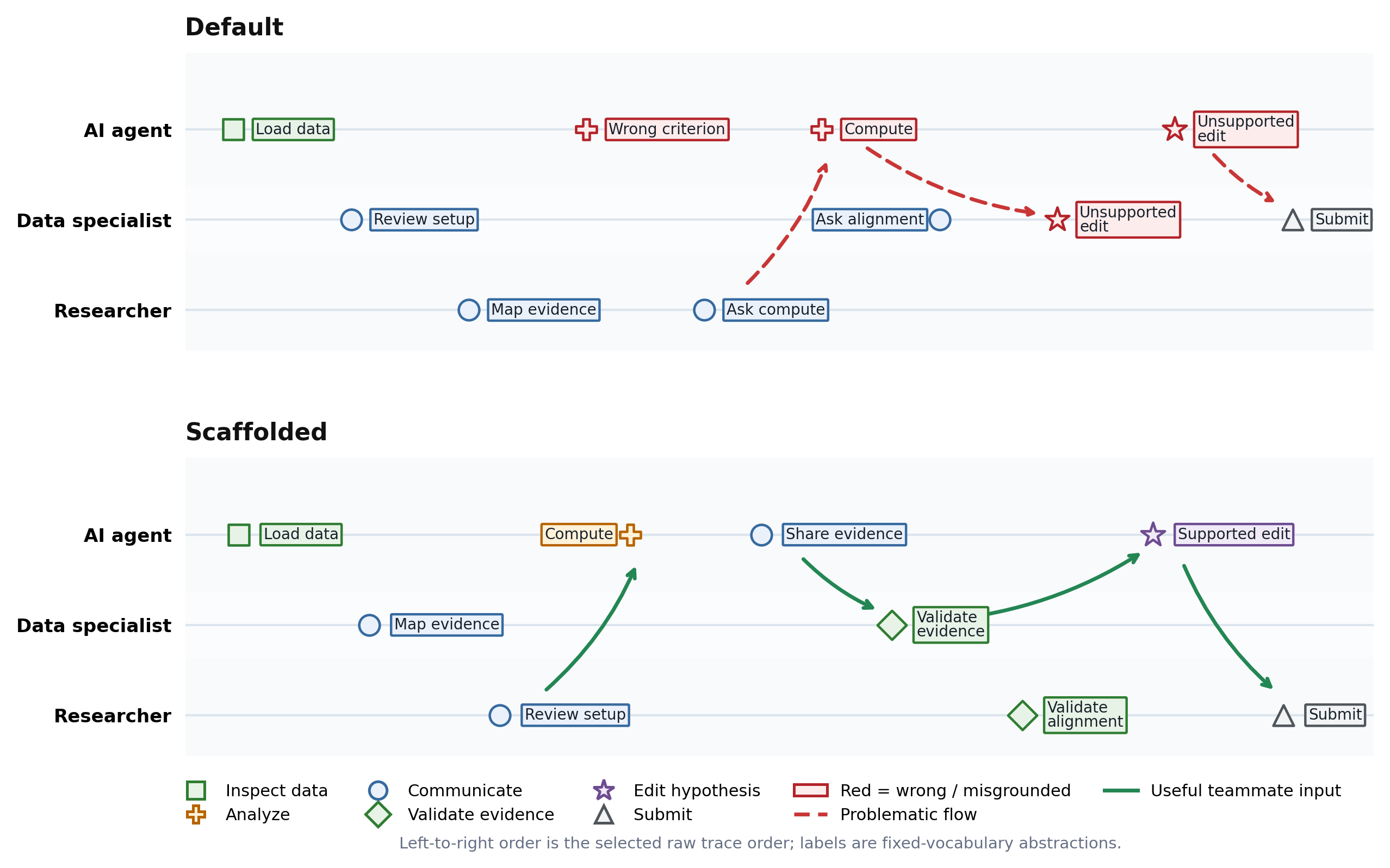}
\caption{Selected-event trace panel}
\end{subfigure}
\caption{Selected trace pair illustrating routed checks in a three-person task. In the Default trace, input from both data-analysis and researcher collaborators is present, but a rolling-window criterion still reaches the submitted hypothesis. In the Scaffolded trace, the team routes variable mapping, evidence validation, and answer--query alignment through designated gate owners before editing.}
\label{fig:qualitative-routed-checks}
\end{figure*}

\subsection{Default Teams Show Process-Loss Patterns}
Default collaborative teams do not improve on the single-agent baseline, and the three-person Default-DR team performs worst. The single-agent baseline reaches $0.71$ Performance; Default-D and Default-R are slightly lower ($0.69$ and $0.68$), and Default-DR falls to $0.63$. The largest drop occurs when both profiled collaborators are present, a pattern consistent with process loss as more distinct expertise must be coordinated.

The drop is not explained by lower activity. Default-DR has more human work and team messages than the single-profile Default teams, yet lower mean Performance; the teams interact more but produce a less-supported hypothesis. Default teams also show lower Hypothesis Support: $S_{\mathrm{hyp}}$ falls from $0.28$ for Single-agent to $0.18$--$0.19$ for profiled Default teams.

\subsection{Scaffolded Teams Raise Mean Performance Across Team Compositions}
Scaffolded teams show a positive mean Performance direction relative to matched Default teams, with the largest and clearest gain in DR ($+0.13$). The D and R gains are smaller ($+0.03$ and $+0.05$) relative to the standard errors, so the main performance pattern is concentrated in the three-person composition, the same composition that had the largest Default drop. Because profiles and models are held fixed across these variants, the main design difference between matched Default and Scaffolded teams is the collaboration structure.

The main process shift is in how initiative is distributed. From Default-DR to Scaffolded-DR, $W_{\mathrm{total}}$ changes only from $7.6$ to $7.9$, while $W_{\mathrm{human}}$ rises from $1.6$ to $2.2$. Across profiles, $H_{\mathrm{init,norm}}$ increases by $+0.31$ to $+0.43$. In DR, Scaffolded also has higher means on Workflow Coverage, Hypothesis Support, and Profile Alignment. Total work is roughly unchanged; the teams distribute it differently.

\subsection{Diagnostic Variants Suggest Complementarity}
\label{sec:diagnostic-variants}
Table~\ref{tab:appendix-diagnostic-results} places the diagnostic variants within each collaborator profile, with Default and Scaffolded repeated as shaded anchors. Shared group memory only removes the gates; the preassigned-gates variant fixes owners by action type. Because neither variant includes the team-led ownership step used in Scaffolded, these comparisons should be read as diagnostic decompositions rather than clean ablations.

Shared group memory only raises $H_{\mathrm{init,norm}}$ by $+0.27$ to $+0.40$ and Team Messages by roughly $+5$ to $+9$ across profiles. Its initiative-entropy values are close to those of Scaffolded, suggesting that the shared-memory component accounts for most of the initiative-distribution shift. However, initiative and communication alone are not sufficient: in R teams, shared group memory only increases both while mean Performance falls from $0.68$ to $0.64$.

Preassigned simulated HITL gates align more directly with Hypothesis Support, raising mean $S_{\mathrm{hyp}}$ in all three profiles, with the largest increases in D and R. In the three-person DR team, neither component alone matches the full Scaffolded Performance mean (Table~\ref{tab:appendix-diagnostic-results}). This is consistent with complementarity: shared group memory gives the team a basis for choosing responsibilities, and simulated HITL gates turn those responsibilities into binding approval requirements. This pattern is clearest in the three-person DR team: Default performs worst there, and the full Scaffolded setting produces the largest improvement.

Figure~\ref{fig:qualitative-routed-checks} grounds this complementarity in one three-person task; Section~\ref{sec:qualitative-analysis} discusses this example alongside two additional trace pairs that show Default process loss and differences in evidential grounding.

\begin{table*}[t]
\caption{Diagnostic variants by collaborator profile. Default and Scaffolded rows are repeated from Table~\ref{tab:results-main} as shaded anchors; the unshaded rows show shared group memory only and preassigned simulated HITL gates. Cells report task-level mean (SE) after averaging across seeds; activity counts are raw, while process and Performance metrics are on $[0,1]$. The preassigned-gates variant fixes gate owners by action type rather than letting the team choose them, so these are diagnostic comparisons rather than symmetric ablations.}
\label{tab:appendix-diagnostic-results}
\centering
\definecolor{anchorrow}{RGB}{236,236,236}
{\setlength\tabcolsep{3.2pt}
\renewcommand{\arraystretch}{1.18}
\resizebox{\textwidth}{!}{%
\begin{tabular}{lrrrrrrrr}
\toprule
\multicolumn{1}{c}{\textbf{Profile / variant}} &
\multicolumn{3}{c}{\textbf{Activity}} &
\multicolumn{4}{c}{\textbf{Process}} &
\multicolumn{1}{c}{\textbf{Outcome}} \\
\cmidrule(r){1-1}
\cmidrule(lr){2-4}
\cmidrule(lr){5-8}
\cmidrule(l){9-9}
\multicolumn{1}{c}{} &
\multicolumn{1}{r}{\small $W_{\mathrm{human}}$} &
\multicolumn{1}{r}{\small $W_{\mathrm{total}}$} &
\multicolumn{1}{r}{\small $M_{\mathrm{team}}$} &
\multicolumn{1}{r}{\small $H_{\mathrm{init,norm}}$} &
\multicolumn{1}{r}{\small $A_{\mathrm{profile}}$} &
\multicolumn{1}{r}{\small $C_{\mathrm{wf}}$} &
\multicolumn{1}{r}{\small $S_{\mathrm{hyp}}$} &
\multicolumn{1}{r}{\small Perf.} \\
\midrule
\multicolumn{9}{l}{\textit{D: data-analysis profile}} \\
\rowcolor{anchorrow} Default &
$0.95$ {\color{stderrgray} $(0.1)$} &
$7.2$ {\color{stderrgray} $(0.3)$} &
$3.5$ {\color{stderrgray} $(0.3)$} &
$0.31$ {\color{stderrgray} $(0.05)$} &
$0.68$ {\color{stderrgray} $(0.04)$} &
$0.65$ {\color{stderrgray} $(0.04)$} &
$0.18$ {\color{stderrgray} $(0.05)$} &
$0.69$ {\color{stderrgray} $(0.06)$} \\
\quad + Shared group memory &
$1.1$ {\color{stderrgray} $(0.1)$} &
$7.2$ {\color{stderrgray} $(0.5)$} &
$8.9$ {\color{stderrgray} $(0.5)$} &
$0.71$ {\color{stderrgray} $(0.04)$} &
$0.67$ {\color{stderrgray} $(0.05)$} &
$0.63$ {\color{stderrgray} $(0.05)$} &
$0.19$ {\color{stderrgray} $(0.05)$} &
$0.73$ {\color{stderrgray} $(0.05)$} \\
\quad + Preassigned gates &
$1.2$ {\color{stderrgray} $(0.08)$} &
$7.2$ {\color{stderrgray} $(0.3)$} &
$4.5$ {\color{stderrgray} $(0.4)$} &
$0.25$ {\color{stderrgray} $(0.04)$} &
$0.67$ {\color{stderrgray} $(0.04)$} &
$0.62$ {\color{stderrgray} $(0.04)$} &
$0.24$ {\color{stderrgray} $(0.05)$} &
$0.74$ {\color{stderrgray} $(0.05)$} \\
\rowcolor{anchorrow} Scaffolded &
$1.2$ {\color{stderrgray} $(0.1)$} &
$6.9$ {\color{stderrgray} $(0.3)$} &
$11$ {\color{stderrgray} $(0.6)$} &
$0.74$ {\color{stderrgray} $(0.04)$} &
$0.69$ {\color{stderrgray} $(0.04)$} &
$0.63$ {\color{stderrgray} $(0.04)$} &
$0.23$ {\color{stderrgray} $(0.06)$} &
$0.72$ {\color{stderrgray} $(0.05)$} \\
\midrule
\multicolumn{9}{l}{\textit{R: researcher profile}} \\
\rowcolor{anchorrow} Default &
$0.62$ {\color{stderrgray} $(0.09)$} &
$7.5$ {\color{stderrgray} $(0.4)$} &
$3.5$ {\color{stderrgray} $(0.3)$} &
$0.37$ {\color{stderrgray} $(0.04)$} &
$0.65$ {\color{stderrgray} $(0.04)$} &
$0.61$ {\color{stderrgray} $(0.04)$} &
$0.18$ {\color{stderrgray} $(0.04)$} &
$0.68$ {\color{stderrgray} $(0.07)$} \\
\quad + Shared group memory &
$1.2$ {\color{stderrgray} $(0.1)$} &
$7.5$ {\color{stderrgray} $(0.3)$} &
$9.5$ {\color{stderrgray} $(0.5)$} &
$0.69$ {\color{stderrgray} $(0.04)$} &
$0.67$ {\color{stderrgray} $(0.05)$} &
$0.62$ {\color{stderrgray} $(0.04)$} &
$0.20$ {\color{stderrgray} $(0.05)$} &
$0.64$ {\color{stderrgray} $(0.05)$} \\
\quad + Preassigned gates &
$0.88$ {\color{stderrgray} $(0.1)$} &
$7.4$ {\color{stderrgray} $(0.3)$} &
$6.1$ {\color{stderrgray} $(0.4)$} &
$0.29$ {\color{stderrgray} $(0.04)$} &
$0.66$ {\color{stderrgray} $(0.04)$} &
$0.62$ {\color{stderrgray} $(0.04)$} &
$0.25$ {\color{stderrgray} $(0.05)$} &
$0.71$ {\color{stderrgray} $(0.05)$} \\
\rowcolor{anchorrow} Scaffolded &
$1.4$ {\color{stderrgray} $(0.3)$} &
$7.5$ {\color{stderrgray} $(0.6)$} &
$12$ {\color{stderrgray} $(0.4)$} &
$0.77$ {\color{stderrgray} $(0.03)$} &
$0.66$ {\color{stderrgray} $(0.04)$} &
$0.61$ {\color{stderrgray} $(0.04)$} &
$0.18$ {\color{stderrgray} $(0.05)$} &
$0.73$ {\color{stderrgray} $(0.06)$} \\
\midrule
\multicolumn{9}{l}{\textit{DR: both profiles}} \\
\rowcolor{anchorrow} Default &
$1.6$ {\color{stderrgray} $(0.2)$} &
$7.6$ {\color{stderrgray} $(0.4)$} &
$6.1$ {\color{stderrgray} $(0.5)$} &
$0.54$ {\color{stderrgray} $(0.03)$} &
$0.66$ {\color{stderrgray} $(0.04)$} &
$0.62$ {\color{stderrgray} $(0.04)$} &
$0.19$ {\color{stderrgray} $(0.06)$} &
$0.63$ {\color{stderrgray} $(0.06)$} \\
\quad + Shared group memory &
$2.7$ {\color{stderrgray} $(0.2)$} &
$8.5$ {\color{stderrgray} $(0.3)$} &
$15$ {\color{stderrgray} $(0.5)$} &
$0.81$ {\color{stderrgray} $(0.02)$} &
$0.69$ {\color{stderrgray} $(0.04)$} &
$0.64$ {\color{stderrgray} $(0.04)$} &
$0.23$ {\color{stderrgray} $(0.06)$} &
$0.69$ {\color{stderrgray} $(0.06)$} \\
\quad + Preassigned gates &
$2.2$ {\color{stderrgray} $(0.3)$} &
$7.5$ {\color{stderrgray} $(0.5)$} &
$6.5$ {\color{stderrgray} $(0.7)$} &
$0.58$ {\color{stderrgray} $(0.02)$} &
$0.68$ {\color{stderrgray} $(0.04)$} &
$0.63$ {\color{stderrgray} $(0.04)$} &
$0.22$ {\color{stderrgray} $(0.06)$} &
$0.64$ {\color{stderrgray} $(0.06)$} \\
\rowcolor{anchorrow} Scaffolded &
$2.2$ {\color{stderrgray} $(0.2)$} &
$7.9$ {\color{stderrgray} $(0.4)$} &
$18$ {\color{stderrgray} $(0.7)$} &
$0.85$ {\color{stderrgray} $(0.02)$} &
$0.70$ {\color{stderrgray} $(0.04)$} &
$0.65$ {\color{stderrgray} $(0.04)$} &
$0.23$ {\color{stderrgray} $(0.06)$} &
$0.76$ {\color{stderrgray} $(0.05)$} \\
\bottomrule
\end{tabular}}}
\end{table*}

\subsection{Qualitative Trace Analysis}
\label{sec:qualitative-analysis}

To connect the aggregate results to trace-level mechanisms, we inspect three trace pairs. Each figure summarizes selected events from the raw traces rather than the full log.

First, Figure~\ref{fig:qualitative-routed-checks} contrasts Default-DR and Scaffolded-DR on a task requiring the second simultaneous decrease of two archaeological signals. In the Default trace, both collaborator lanes are active, but a rolling-window criterion still flows into the editor and the final hypothesis. In the Scaffolded trace, checks precede finalization: the data-analysis collaborator maps and later validates the evidence, the researcher checks setup and answer--query alignment, and the agent computes candidate events before editing. The contrast illustrates the intended mechanism: existing expertise is made operational before finalization, rather than changing the team composition.

Second, Figure~\ref{fig:appendix-copper-peak-panel} compares Default-D and Scaffolded-D traces for a copper-peak task. In the Default trace, the data-analysis collaborator asks for broader evidence and answer--query alignment, but those requests never become a binding check on the analysis criterion: the agent follows a smoothed peak-detection path and records an unsupported 9th-century BCE hypothesis, while the reference points to the earlier 35th-century BCE peak. In the Scaffolded trace for the same task, the collaborator's criterion check precedes comparison and a revised computation before the hypothesis is edited.

Third, Figure~\ref{fig:appendix-pottery-decoration-panel} compares Default and Scaffolded traces for a pottery-decoration task. The Default team finds the relevant signal but anchors the hypothesis to the first elevated period rather than the start of the highest sustained plateau. In the Scaffolded trace, the team computes and validates the later maximum plateau before submission, producing a more directly supported hypothesis, though it still does not explicitly contrast the lower and highest plateaus.

\newcommand{\abstractTaskSeven}{%
\fcolorbox{taskgrid}{tasklight}{%
\begin{minipage}[t]{0.94\linewidth}
\setlength{\fboxsep}{1.5pt}
\footnotesize
\begin{minipage}[t]{0.32\linewidth}
\vspace{0pt}
\datasetstag\\[0.18em]
\activefiletag{time\_series\_data.csv}\\[0.22em]
\inactivefiletag{capital.csv}\\[0.22em]
\inactivefiletag{pollen\_\ldots.csv}\\[0.22em]
\morefiletag
\end{minipage}\hfill
\begin{minipage}[t]{0.62\linewidth}
\vspace{0pt}
\scriptsize
\setlength{\tabcolsep}{3.5pt}
\setlength{\arrayrulewidth}{0.45pt}
\renewcommand{\arraystretch}{1.16}
\arrayrulecolor{taskgrid}
\begin{tabularx}{\linewidth}{|>{\centering\arraybackslash}p{0.17\linewidth}|>{\centering\arraybackslash}p{0.30\linewidth}|>{\centering\arraybackslash}p{0.30\linewidth}|>{\centering\arraybackslash}X|}
\hline
\rowcolor{taskheader}\textbf{CE} & \textbf{CopperGold} & \textbf{Copper\_inter} & \textbf{\ldots} \\
\hline
\rowcolor{taskcell}
-3600 & -0.50 & -0.50 & \ldots \\
\hline
\rowcolor{taskcell}
-3500 & 0.69 & 0.69 & \ldots \\
\hline
\rowcolor{taskcell}
-3400 & 0.69 & 0.69 & \ldots \\
\hline
\rowcolor{taskcell}
\ldots & \ldots & \ldots & \ldots \\
\hline
\end{tabularx}
\arrayrulecolor{black}
\setlength{\arrayrulewidth}{0.4pt}
\end{minipage}
\par\vspace{0.45em}\noindent
{\bfseries Goal:} In which century did copper have its first peak?
\end{minipage}}}

\newcommand{\abstractTaskThirtyOne}{%
\fcolorbox{taskgrid}{tasklight}{%
\begin{minipage}[t]{0.94\linewidth}
\setlength{\fboxsep}{1.5pt}
\footnotesize
\begin{minipage}[t]{0.32\linewidth}
\vspace{0pt}
\datasetstag\\[0.18em]
\activefiletag{time\_series\_data.csv}\\[0.22em]
\inactivefiletag{capital.csv}\\[0.22em]
\inactivefiletag{pollen\_\ldots.csv}\\[0.22em]
\morefiletag
\end{minipage}\hfill
\begin{minipage}[t]{0.62\linewidth}
\vspace{0pt}
\scriptsize
\setlength{\tabcolsep}{3.5pt}
\setlength{\arrayrulewidth}{0.45pt}
\renewcommand{\arraystretch}{1.16}
\arrayrulecolor{taskgrid}
\begin{tabularx}{\linewidth}{|>{\centering\arraybackslash}p{0.17\linewidth}|>{\centering\arraybackslash}p{0.30\linewidth}|>{\centering\arraybackslash}p{0.30\linewidth}|>{\centering\arraybackslash}X|}
\hline
\rowcolor{taskheader}\textbf{CE} & \textbf{PotteryDec.} & \textbf{Decor\_inter} & \textbf{\ldots} \\
\hline
\rowcolor{taskcell}
-4100 & -1.92 & -1.9200 & \ldots \\
\hline
\rowcolor{taskcell}
-4000 & -0.11 & -0.1100 & \ldots \\
\hline
\rowcolor{taskcell}
-3900 & -0.26 & -0.2600 & \ldots \\
\hline
\rowcolor{taskcell}
\ldots & \ldots & \ldots & \ldots \\
\hline
\end{tabularx}
\arrayrulecolor{black}
\setlength{\arrayrulewidth}{0.4pt}
\end{minipage}
\par\vspace{0.45em}\noindent
{\bfseries Goal:} In which century does Diversity in Pottery Decoration begin to show its highest sustained values?
\end{minipage}}}

\section{Concluding Discussion}

Taken together, the results show why simulated shared-workspace teams are useful as a controlled layer between single-agent benchmarks and live human studies. The testbed lets us compare the same tasks, models, interfaces, and collaborator profiles while changing the collaboration structure. In this setting, team outcomes depend not only on available expertise, but on whether the interaction gives that expertise a path into the submitted hypothesis.

This matters for how the field builds and evaluates collaborative agents. Training and inference pipelines need to attend to the collaborative processes within trajectories (responsibility assignment, evidence handoff, and review routing), because these mechanisms determine whether complementary expertise actually reaches the team's final product. The group-process literature offers a starting point: responsibility assignment, transactive memory, and structured review are well-studied interventions for human teams, and our simulation results show that adaptations of these ideas shift coordination patterns in simulated human--AI teams as well.

\paragraph{Limitations and Future Work.}
The scaffolded setting we evaluate is a first hand-designed probe, not a final account of how human--AI teams should coordinate. Simulated collaborators remain approximations of real people: they build on prior human-agent and active-user-simulation frameworks \citep{shao2024cogym,nathani2026pare}, but may show simulator-specific artifacts or underrepresent variation in proactiveness and strategy. Recent work on scientific deep-research agents has begun collecting expert feedback on intermediate actions, showing that user-preferred actions are highly user-specific and better predicted from interaction histories \citep{balepur2026dracula}. Our setting needs analogous data for longer shared-workspace, multi-participant collaboration: real human traces can validate these collaborators and support simulator populations with varied expertise, proactiveness, and work strategies \citep{mehri2026distributional,chopra2026personas}. The goal is not to replace real-user evaluation, but to use simulation to generate sharper hypotheses and candidate interventions before human studies. Future work can use such simulators to search over richer collaboration policies that adaptively request expertise, assign responsibilities, route checks, or relax structure as teams converge. Long multi-participant trajectories can be sensitive to early choices \citep{laban2025lost}; additional seeds, larger task suites, and variance-reduction sampling would support future causal claims. The archaeology subset provides domain-specific interpretation challenges, but future work should test generalization. Finally, with validated simulators and richer traces, future work can move from evaluating fixed scaffolds to optimizing collaboration policies using reinforcement learning \citep{wu2025collabllm,zhou2025sweetrl}.

\section*{Impact Statement}
This work studies collaboration structure in simulated human--AI data-analysis teams. The scaffolds we evaluate (shared group memory and simulated HITL gates) preserve simulated approval points corresponding to human roles rather than treating collaboration as full automation. These findings should be validated with real users before high-stakes deployment. More broadly, collaborative AI systems should preserve human agency and augment human judgment rather than replace it.

\clearpage

\bibliography{draft}

@article{vaccaro2024humanai,
  author = {Vaccaro, Michelle and Almaatouq, Abdullah and Malone, Thomas W.},
  title = {When Combinations of Humans and {AI} Are Useful: A Systematic Review and Meta-Analysis},
  journal = {Nature Human Behaviour},
  year = {2024},
  volume = {8},
  pages = {2293--2303},
  Xdoi = {10.1038/s41562-024-02024-1},
  xURL = {https://doi.org/10.1038/s41562-024-02024-1}
}

@inproceedings{bansal2021whole,
  author       = {Gagan Bansal and others},
  Xauthor       = {Gagan Bansal and
                  Tongshuang Wu and
                  Joyce Zhou and
                  Raymond Fok and
                  Besmira Nushi and
                  Ece Kamar and
                  Marco T{\'{u}}lio Ribeiro and
                  Daniel S. Weld},
  Xeditor       = {Yoshifumi Kitamura and
                  Aaron Quigley and
                  Katherine Isbister and
                  Takeo Igarashi and
                  Pernille Bj{\o}rn and
                  Steven Mark Drucker},
  title        = {Does the Whole Exceed its Parts? The Effect of {AI} Explanations on
                  Complementary Team Performance},
  booktitle    = {Proceedings of the Conference on Human Factors in Computing Systems (CHI)},
  Xbooktitle    = {{CHI} '21: {CHI} Conference on Human Factors in Computing Systems,
                  Virtual Event / Yokohama, Japan, May 8-13, 2021},
  Xpages        = {81:1--81:16},
  Xpublisher    = {{ACM}},
  year         = {2021},
  Xurl          = {https://doi.org/10.1145/3411764.3445717},
  Xdoi          = {10.1145/3411764.3445717},
  Xtimestamp    = {Sun, 02 Nov 2025 21:27:18 +0100},
  Xbiburl       = {https://dblp.org/rec/conf/chi/BansalWZFNKRW21.bib},
  Xbibsource    = {dblp computer science bibliography, https://dblp.org}
}

@techreport{agarwal2023radiology,
  author = {Agarwal, Nikhil and Moehring, Alex and Rajpurkar, Pranav and Salz, Tobias},
  title = {Combining Human Expertise with Artificial Intelligence: Experimental Evidence from Radiology},
  institution = {National Bureau of Economic Research},
  Xtype = {Working Paper},
  number = {31422},
  year = {2023},
  Xmonth = jul,
  Xdoi = {10.3386/w31422},
  xURL = {https://www.nber.org/papers/w31422}
}

@article{yu2024heterogeneity,
  author = {Yu, Feiyang and Moehring, Alex and Banerjee, Oishi and Salz, Tobias and Agarwal, Nikhil and Rajpurkar, Pranav},
  title = {Heterogeneity and Predictors of the Effects of {AI} Assistance on Radiologists},
  journal = {Nature Medicine},
  year = {2024},
  volume = {30},
  pages = {837--849},
  Xdoi = {10.1038/s41591-024-02850-w},
  xURL = {https://doi.org/10.1038/s41591-024-02850-w}
}

@book{steiner1972group,
  author = {Steiner, Ivan D.},
  title = {Group Process and Productivity},
  publisher = {Academic Press},
  Xaddress = {New York, NY},
  year = {1972}
}

@article{malone1994coordination,
  author       = {Thomas W. Malone and
                  Kevin Crowston},
  title        = {The Interdisciplinary Study of Coordination},
  journal      = {{ACM} Comput. Surv.},
  volume       = {26},
  number       = {1},
  pages        = {87--119},
  year         = {1994},
  Xurl          = {https://doi.org/10.1145/174666.174668},
  Xdoi          = {10.1145/174666.174668},
  Xtimestamp    = {Fri, 30 Nov 2018 12:48:46 +0100},
  Xbiburl       = {https://dblp.org/rec/journals/csur/MaloneC94.bib},
  Xbibsource    = {dblp computer science bibliography, https://dblp.org}
}

@article{heath2000coordination,
  author = {Heath, Chip and Staudenmayer, Nancy},
  title = {Coordination Neglect: How Lay Theories of Organizing Complicate Coordination in Organizations},
  journal = {Research in Organizational Behavior},
  year = {2000},
  volume = {22},
  pages = {153--191},
  Xdoi = {10.1016/S0191-3085(00)22005-4},
  xURL = {https://doi.org/10.1016/S0191-3085(00)22005-4}
}

@article{stasser1985pooling,
  author = {Stasser, Garold and Titus, William},
  title = {Pooling of Unshared Information in Group Decision Making: Biased Information Sampling During Discussion},
  journal = {Journal of Personality and Social Psychology},
  year = {1985},
  volume = {48},
  number = {6},
  pages = {1467--1478},
  Xdoi = {10.1037/0022-3514.48.6.1467},
  xURL = {https://doi.org/10.1037/0022-3514.48.6.1467}
}

@article{gutwin2002workspace,
  author       = {Carl Gutwin and
                  Saul Greenberg},
  title        = {A Descriptive Framework of Workspace Awareness for Real-Time Groupware},
  journal      = {Comput. Support. Cooperative Work.},
  volume       = {11},
  number       = {3-4},
  pages        = {411--446},
  year         = {2002},
  Xurl          = {https://doi.org/10.1023/A:1021271517844},
  Xdoi          = {10.1023/A:1021271517844},
  Xtimestamp    = {Fri, 05 Jun 2020 14:00:07 +0200},
  Xbiburl       = {https://dblp.org/rec/journals/cscw/GutwinG02.bib},
  Xbibsource    = {dblp computer science bibliography, https://dblp.org}
}

@article{shao2024cogym,
  author       = {Yijia Shao and
                  Vinay Samuel and
                  Yucheng Jiang and
                  John Yang and
                  Diyi Yang},
  title        = {Collaborative {Gym}: {A} Framework for Enabling and Evaluating Human-Agent Collaboration},
  journal      = {CoRR},
  volume       = {abs/2412.15701},
  year         = {2024},
  Xurl          = {https://doi.org/10.48550/arXiv.2412.15701},
  Xdoi          = {10.48550/ARXIV.2412.15701},
  Xeprinttype   = {arXiv},
  Xeprint       = {2412.15701},
  Xtimestamp    = {Mon, 02 Jun 2025 21:06:30 +0200},
  Xbiburl       = {https://dblp.org/rec/journals/corr/abs-2412-15701.bib},
  Xbibsource    = {dblp computer science bibliography, https://dblp.org}
}

@inproceedings{sun2025collabovercooked,
  author       = {Haochen Sun and others},
  Xauthor       = {Haochen Sun and
                  Shuwen Zhang and
                  Lujie Niu and
                  Lei Ren and
                  Hao Xu and
                  Hao Fu and
                  Fangkun Zhao and
                  Caixia Yuan and
                  Xiaojie Wang},
  Xeditor       = {Christos Christodoulopoulos and
                  Tanmoy Chakraborty and
                  Carolyn Rose and
                  Violet Peng},
  title        = {{Collab-Overcooked}: Benchmarking and Evaluating Large Language Models as Collaborative Agents},
  booktitle    = {Proceedings of the Conference on Empirical Methods in Natural Language Processing (EMNLP)},
  Xbooktitle    = {Proceedings of the 2025 Conference on Empirical Methods in Natural
                  Language Processing, {EMNLP} 2025, Suzhou, China, November 4-9, 2025},
  Xpages        = {4922--4951},
  Xpublisher    = {Association for Computational Linguistics},
  year         = {2025},
  Xurl          = {https://doi.org/10.18653/v1/2025.emnlp-main.249},
  Xdoi          = {10.18653/V1/2025.EMNLP-MAIN.249},
  Xtimestamp    = {Wed, 10 Jun 2026 11:17:00 +0200},
  Xbiburl       = {https://dblp.org/rec/conf/emnlp/SunZNRXFZYW25.bib},
  Xbibsource    = {dblp computer science bibliography, https://dblp.org}
}

@article{nathani2026pare,
  author       = {Deepak Nathani and others},
  Xauthor       = {Deepak Nathani and
                  Cheng Zhang and
                  Chang Huan and
                  Jiaming Shan and
                  Yinfei Yang and
                  Alkesh Patel and
                  Zhe Gan and
                  William Yang Wang and
                  Michael Saxon and
                  Xin Eric Wang},
  title        = {Proactive Agent Research Environment: Simulating Active Users to Evaluate
                  Proactive Assistants},
  journal      = {CoRR},
  volume       = {abs/2604.00842},
  year         = {2026},
  Xurl          = {https://doi.org/10.48550/arXiv.2604.00842},
  Xdoi          = {10.48550/ARXIV.2604.00842},
  Xeprinttype   = {arXiv},
  Xeprint       = {2604.00842},
  Xtimestamp    = {Thu, 07 May 2026 09:02:17 +0200},
  Xbiburl       = {https://dblp.org/rec/journals/corr/abs-2604-00842.bib},
  Xbibsource    = {dblp computer science bibliography, https://dblp.org}
}

@inproceedings{gonzalezpumariega2025robotouille,
  author       = {Gonzalo Gonzalez{-}Pumariega and
                  Leong Su Yean and
                  Neha Sunkara and
                  Sanjiban Choudhury},
  title        = {Robotouille: An Asynchronous Planning Benchmark for {LLM} Agents},
  booktitle    = {Proceedings of the International Conference on Learning Representations (ICLR)},
  Xbooktitle    = {The Thirteenth International Conference on Learning Representations,
                  {ICLR} 2025, Singapore, April 24-28, 2025},
  Xpublisher    = {OpenReview.net},
  year         = {2025},
  Xurl          = {https://openreview.net/forum?id=OhUoTMxFIH},
  Xtimestamp    = {Thu, 15 May 2025 17:19:05 +0200},
  Xbiburl       = {https://dblp.org/rec/conf/iclr/Gonzalez-Pumariega25.bib},
  Xbibsource    = {dblp computer science bibliography, https://dblp.org}
}

@inproceedings{lin2024asynchow,
  author       = {Fangru Lin and
                  Emanuele La Malfa and
                  Valentin Hofmann and
                  Elle Michelle Yang and
                  Anthony G. Cohn and
                  Janet B. Pierrehumbert},
  Xeditor       = {Ruslan Salakhutdinov and
                  Zico Kolter and
                  Katherine A. Heller and
                  Adrian Weller and
                  Nuria Oliver and
                  Jonathan Scarlett and
                  Felix Berkenkamp},
  title        = {Graph-enhanced Large Language Models in Asynchronous Plan Reasoning},
  booktitle    = {Proceedings of the International Conference on Machine Learning (ICML)},
  Xbooktitle    = {Forty-first International Conference on Machine Learning, {ICML} 2024,
                  Vienna, Austria, July 21-27, 2024},
  Xseries       = {Proceedings of Machine Learning Research},
  Xvolume       = {235},
  Xpages        = {30108--30134},
  Xpublisher    = {{PMLR} / OpenReview.net},
  year         = {2024},
  Xurl          = {https://proceedings.mlr.press/v235/lin24k.html},
  Xtimestamp    = {Mon, 09 Feb 2026 17:23:53 +0100},
  Xbiburl       = {https://dblp.org/rec/conf/icml/LinMHY0P24.bib},
  Xbibsource    = {dblp computer science bibliography, https://dblp.org}
}

@article{paracook2025,
  author       = {Shiqi Zhang and others},
  Xauthor       = {Shiqi Zhang and
                  Xinbei Ma and
                  Yunqing Xu and
                  Zouying Cao and
                  Pengrui Lu and
                  Haobo Yuan and
                  Tiancheng Shen and
                  Zhuosheng Zhang and
                  Hai Zhao and
                  Ming{-}Hsuan Yang},
  title        = {{ParaCook}: On Time-Efficient Planning for Multi-Agent Systems},
  journal      = {CoRR},
  volume       = {abs/2510.11608},
  year         = {2025},
  Xurl          = {https://doi.org/10.48550/arXiv.2510.11608},
  Xdoi          = {10.48550/ARXIV.2510.11608},
  Xeprinttype   = {arXiv},
  Xeprint       = {2510.11608},
  Xtimestamp    = {Wed, 12 Nov 2025 14:27:37 +0100},
  Xbiburl       = {https://dblp.org/rec/journals/corr/abs-2510-11608.bib},
  Xbibsource    = {dblp computer science bibliography, https://dblp.org}
}

@article{masters2025manageragentgym,
  author       = {Charlie Masters and others},
  Xauthor       = {Charlie Masters and
                  Advaith Vellanki and
                  Jiangbo Shangguan and
                  Bart Kultys and
                  Jonathan Gilmore and
                  Alastair Moore and
                  Stefano V. Albrecht},
  title        = {Orchestrating Human-{AI} Teams: The Manager Agent as a Unifying Research Challenge},
  journal      = {CoRR},
  volume       = {abs/2510.02557},
  year         = {2025},
  Xurl          = {https://doi.org/10.48550/arXiv.2510.02557},
  Xdoi          = {10.48550/ARXIV.2510.02557},
  Xeprinttype   = {arXiv},
  Xeprint       = {2510.02557},
  Xtimestamp    = {Sun, 09 Nov 2025 15:58:16 +0100},
  Xbiburl       = {https://dblp.org/rec/journals/corr/abs-2510-02557.bib},
  Xbibsource    = {dblp computer science bibliography, https://dblp.org}
}

@inproceedings{majumder2024discoverybench,
  author       = {Bodhisattwa Prasad Majumder and others},
  Xauthor       = {Bodhisattwa Prasad Majumder and
                  Harshit Surana and
                  Dhruv Agarwal and
                  Bhavana Dalvi Mishra and
                  Abhijeetsingh Meena and
                  Aryan Prakhar and
                  Tirth Vora and
                  Tushar Khot and
                  Ashish Sabharwal and
                  Peter Clark},
  title        = {{DiscoveryBench}: Towards Data-Driven Discovery with Large Language Models},
  booktitle    = {Proceedings of the International Conference on Learning Representations (ICLR)},
  Xbooktitle    = {The Thirteenth International Conference on Learning Representations,
                  {ICLR} 2025, Singapore, April 24-28, 2025},
  Xpublisher    = {OpenReview.net},
  year         = {2025},
  Xurl          = {https://openreview.net/forum?id=vyflgpwfJW},
  Xtimestamp    = {Thu, 15 May 2025 17:19:05 +0200},
  Xbiburl       = {https://dblp.org/rec/conf/iclr/MajumderS0MMPVK25.bib},
  Xbibsource    = {dblp computer science bibliography, https://dblp.org}
}

@article{laban2025lost,
  author       = {Philippe Laban and
                  Hiroaki Hayashi and
                  Yingbo Zhou and
                  Jennifer Neville},
  title        = {{LLMs} Get Lost in Multi-Turn Conversation},
  journal      = {CoRR},
  volume       = {abs/2505.06120},
  year         = {2025},
  Xurl          = {https://doi.org/10.48550/arXiv.2505.06120},
  Xdoi          = {10.48550/ARXIV.2505.06120},
  Xeprinttype   = {arXiv},
  Xeprint       = {2505.06120},
  Xtimestamp    = {Thu, 05 Feb 2026 17:35:56 +0100},
  Xbiburl       = {https://dblp.org/rec/journals/corr/abs-2505-06120.bib},
  Xbibsource    = {dblp computer science bibliography, https://dblp.org}
}

@inproceedings{yao2023react,
  author       = {Shunyu Yao and others},
  Xauthor       = {Shunyu Yao and
                  Jeffrey Zhao and
                  Dian Yu and
                  Nan Du and
                  Izhak Shafran and
                  Karthik R. Narasimhan and
                  Yuan Cao},
  title        = {{ReAct}: Synergizing Reasoning and Acting in Language Models},
  booktitle    = {Proceedings of the International Conference on Learning Representations (ICLR)},
  Xbooktitle    = {The Eleventh International Conference on Learning Representations,
                  {ICLR} 2023, Kigali, Rwanda, May 1-5, 2023},
  Xpublisher    = {OpenReview.net},
  year         = {2023},
  Xurl          = {https://openreview.net/forum?id=WE\_vluYUL-X},
  Xtimestamp    = {Fri, 19 Dec 2025 20:56:24 +0100},
  Xbiburl       = {https://dblp.org/rec/conf/iclr/YaoZYDSN023.bib},
  Xbibsource    = {dblp computer science bibliography, https://dblp.org}
}

@inproceedings{wu2025collabllm,
  author       = {Shirley Wu and others},
  Xauthor       = {Shirley Wu and
                  Michel Galley and
                  Baolin Peng and
                  Hao Cheng and
                  Gavin Li and
                  Yao Dou and
                  Weixin Cai and
                  James Zou and
                  Jure Leskovec and
                  Jianfeng Gao},
  Xeditor       = {Aarti Singh and
                  Maryam Fazel and
                  Daniel Hsu and
                  Simon Lacoste{-}Julien and
                  Felix Berkenkamp and
                  Tegan Maharaj and
                  Kiri Wagstaff and
                  Jerry Zhu},
  title        = {{CollabLLM}: From Passive Responders to Active Collaborators},
  booktitle    = {Proceedings of the International Conference on Machine Learning (ICML)},
  Xbooktitle    = {Forty-second International Conference on Machine Learning, {ICML}
                  2025, Vancouver, BC, Canada, July 13-19, 2025},
  Xseries       = {Proceedings of Machine Learning Research},
  Xvolume       = {267},
  Xpublisher    = {{PMLR} / OpenReview.net},
  year         = {2025},
  Xurl          = {https://proceedings.mlr.press/v267/wu25i.html},
  Xtimestamp    = {Wed, 04 Feb 2026 17:22:45 +0100},
  Xbiburl       = {https://dblp.org/rec/conf/icml/WuGP0LDC0L025.bib},
  Xbibsource    = {dblp computer science bibliography, https://dblp.org}
}

@article{zhou2025sweetrl,
  author       = {Yifei Zhou and others},
  Xauthor       = {Yifei Zhou and
                  Song Jiang and
                  Yuandong Tian and
                  Jason Weston and
                  Sergey Levine and
                  Sainbayar Sukhbaatar and
                  Xian Li},
  title        = {{SWEET-RL:} Training Multi-Turn {LLM} Agents on Collaborative Reasoning
                  Tasks},
  journal      = {CoRR},
  volume       = {abs/2503.15478},
  year         = {2025},
  Xurl          = {https://doi.org/10.48550/arXiv.2503.15478},
  Xdoi          = {10.48550/ARXIV.2503.15478},
  Xeprinttype   = {arXiv},
  Xeprint       = {2503.15478},
  Xtimestamp    = {Wed, 23 Apr 2025 07:56:29 +0200},
  Xbiburl       = {https://dblp.org/rec/journals/corr/abs-2503-15478.bib},
  Xbibsource    = {dblp computer science bibliography, https://dblp.org}
}

@inproceedings{park2023generativeagents,
  author       = {Joon Sung Park and
                  Joseph C. O'Brien and
                  Carrie Jun Cai and
                  Meredith Ringel Morris and
                  Percy Liang and
                  Michael S. Bernstein},
  Xeditor       = {Sean Follmer and
                  Jeff Han and
                  J{\"{u}}rgen Steimle and
                  Nathalie Henry Riche},
  title        = {Generative Agents: Interactive Simulacra of Human Behavior},
  booktitle    = {Proceedings of the ACM Symposium on User Interface Software and Technology (UIST)},
  Xbooktitle    = {Proceedings of the 36th Annual {ACM} Symposium on User Interface Software
                  and Technology, {UIST} 2023, San Francisco, CA, USA, 29 October 2023-
                  1 November 2023},
  Xpages        = {2:1--2:22},
  Xpublisher    = {{ACM}},
  year         = {2023},
  Xurl          = {https://doi.org/10.1145/3586183.3606763},
  Xdoi          = {10.1145/3586183.3606763},
  Xtimestamp    = {Sun, 03 May 2026 14:26:44 +0200},
  Xbiburl       = {https://dblp.org/rec/conf/uist/ParkOCMLB23.bib},
  Xbibsource    = {dblp computer science bibliography, https://dblp.org}
}

@inproceedings{zhou2024sotopia,
  author       = {Xuhui Zhou and others},
  Xauthor       = {Xuhui Zhou and
                  Hao Zhu and
                  Leena Mathur and
                  Ruohong Zhang and
                  Haofei Yu and
                  Zhengyang Qi and
                  Louis{-}Philippe Morency and
                  Yonatan Bisk and
                  Daniel Fried and
                  Graham Neubig and
                  Maarten Sap},
  title        = {{SOTOPIA:} Interactive Evaluation for Social Intelligence in Language
                  Agents},
  booktitle    = {Proceedings of the International Conference on Learning Representations (ICLR)},
  Xbooktitle    = {The Twelfth International Conference on Learning Representations,
                  {ICLR} 2024, Vienna, Austria, May 7-11, 2024},
  Xpublisher    = {OpenReview.net},
  year         = {2024},
  Xurl          = {https://openreview.net/forum?id=mM7VurbA4r},
  Xtimestamp    = {Mon, 29 Jul 2024 17:17:48 +0200},
  Xbiburl       = {https://dblp.org/rec/conf/iclr/Zhou0MZYQMBFNS24.bib},
  Xbibsource    = {dblp computer science bibliography, https://dblp.org}
}

@incollection{wegner1987transactive,
  author = {Wegner, Daniel M.},
  title = {Transactive Memory: A Contemporary Analysis of the Group Mind},
  booktitle = {Theories of Group Behavior},
  Xeditor = {Mullen, Brian and Goethals, George R.},
  Xpages = {185--208},
  publisher = {Springer},
  Xaddress = {New York, NY},
  Xseries = {Springer Series in Social Psychology},
  year = {1987},
  Xdoi = {10.1007/978-1-4612-4634-3_9}
}

@article{lewis2003tms,
  author = {Lewis, Kyle},
  title = {Measuring Transactive Memory Systems in the Field: Scale Development and Validation},
  journal = {Journal of Applied Psychology},
  volume = {88},
  number = {4},
  pages = {587--604},
  year = {2003},
  Xdoi = {10.1037/0021-9010.88.4.587}
}

@article{argote2012tms,
  author = {Argote, Linda and Ren, Yuqing},
  title = {Transactive Memory Systems: A Microfoundation of Dynamic Capabilities},
  journal = {Journal of Management Studies},
  volume = {49},
  number = {8},
  pages = {1375--1382},
  year = {2012},
  Xdoi = {10.1111/j.1467-6486.2012.01077.x}
}

@article{lewis2011tms,
  author       = {Kyle Lewis and
                  Benjamin Herndon},
  title        = {Transactive Memory Systems: Current Issues and Future Research Directions},
  journal      = {Organ. Sci.},
  volume       = {22},
  number       = {5},
  pages        = {1254--1265},
  year         = {2011},
  Xurl          = {https://doi.org/10.1287/orsc.1110.0647},
  Xdoi          = {10.1287/ORSC.1110.0647},
  Xtimestamp    = {Thu, 16 Jul 2020 16:38:24 +0200},
  Xbiburl       = {https://dblp.org/rec/journals/orgsci/LewisH11.bib},
  Xbibsource    = {dblp computer science bibliography, https://dblp.org}
}

@incollection{moreland1999transactive,
  author = {Moreland, Richard L.},
  title = {Transactive Memory: Learning Who Knows What in Work Groups and Organizations},
  booktitle = {Shared Cognition in Organizations: The Management of Knowledge},
  Xeditor = {Thompson, Leigh and Levine, John M. and Messick, David M.},
  Xpages = {3--31},
  publisher = {Lawrence Erlbaum Associates},
  Xaddress = {Mahwah, NJ},
  year = {1999},
  Xdoi = {10.4324/9781410603227-1}
}

@article{mathieu2000sharedmentalmodels,
  author = {Mathieu, John E. and Heffner, Tonia S. and Goodwin, Gerald F. and Salas, Eduardo and Cannon-Bowers, Janis A.},
  title = {The Influence of Shared Mental Models on Team Process and Performance},
  journal = {Journal of Applied Psychology},
  volume = {85},
  number = {2},
  pages = {273--283},
  year = {2000},
  Xdoi = {10.1037/0021-9010.85.2.273}
}

@article{andrews2022sharedmentalmodels,
  author = {Andrews, R. W. and Lilly, J. M. and Srivastava, Divya K. and Feigh, Karen M.},
  title = {The Role of Shared Mental Models in Human-{AI} Teams: A Theoretical Review},
  journal = {Theoretical Issues in Ergonomics Science},
  volume = {24},
  number = {2},
  pages = {129--175},
  year = {2023},
  Xdoi = {10.1080/1463922X.2022.2061080},
  xURL = {https://doi.org/10.1080/1463922X.2022.2061080}
}

@article{fischer2013script,
  author = {Fischer, Frank and Kollar, Ingo and Stegmann, Karsten and Wecker, Christof},
  title = {Toward a Script Theory of Guidance in Computer-Supported Collaborative Learning},
  journal = {Educational Psychologist},
  volume = {48},
  number = {1},
  pages = {56--66},
  year = {2013},
  Xdoi = {10.1080/00461520.2012.748005}
}

@article{kollar2006collaboration,
  author = {Kollar, Ingo and Fischer, Frank and Hesse, Friedrich W.},
  title = {Collaboration Scripts: A Conceptual Analysis},
  journal = {Educational Psychology Review},
  volume = {18},
  pages = {159--185},
  year = {2006},
  Xdoi = {10.1007/s10648-006-9007-2}
}

@inproceedings{horvitz1999mixedinitiative,
  author       = {Eric Horvitz},
  Xeditor       = {Marian G. Williams and
                  Mark W. Altom},
  title        = {Principles of Mixed-Initiative User Interfaces},
  booktitle    = {Proceedings of the Conference on Human Factors in Computing Systems (CHI)},
  Xbooktitle    = {Proceeding of the {CHI} '99 Conference on Human Factors in Computing
                  Systems: The {CHI} is the Limit, Pittsburgh, PA, USA, May 15-20, 1999},
  Xpages        = {159--166},
  Xpublisher    = {{ACM}},
  year         = {1999},
  Xurl          = {https://doi.org/10.1145/302979.303030},
  Xdoi          = {10.1145/302979.303030},
  Xtimestamp    = {Fri, 12 Mar 2021 15:27:48 +0100},
  Xbiburl       = {https://dblp.org/rec/conf/chi/Horvitz99.bib},
  Xbibsource    = {dblp computer science bibliography, https://dblp.org}
}

@article{parasuraman2000automation,
  author       = {Raja Parasuraman and
                  Thomas B. Sheridan and
                  Christopher D. Wickens},
  title        = {A model for types and levels of human interaction with automation},
  journal      = {{IEEE} Trans. Syst. Man Cybern. Part {A}},
  volume       = {30},
  number       = {3},
  pages        = {286--297},
  year         = {2000},
  Xurl          = {https://doi.org/10.1109/3468.844354},
  Xdoi          = {10.1109/3468.844354},
  Xtimestamp    = {Sat, 30 May 2020 19:50:09 +0200},
  Xbiburl       = {https://dblp.org/rec/journals/tsmc/ParasuramanSW00.bib},
  Xbibsource    = {dblp computer science bibliography, https://dblp.org}
}

@inproceedings{amershi2019guidelines,
  author       = {Saleema Amershi and others},
  Xauthor       = {Saleema Amershi and
                  Daniel S. Weld and
                  Mihaela Vorvoreanu and
                  Adam Fourney and
                  Besmira Nushi and
                  Penny Collisson and
                  Jina Suh and
                  Shamsi T. Iqbal and
                  Paul N. Bennett and
                  Kori Inkpen and
                  Jaime Teevan and
                  Ruth Kikin{-}Gil and
                  Eric Horvitz},
  Xeditor       = {Stephen A. Brewster and
                  Geraldine Fitzpatrick and
                  Anna L. Cox and
                  Vassilis Kostakos},
  title        = {Guidelines for Human-{AI} Interaction},
  booktitle    = {Proceedings of the Conference on Human Factors in Computing Systems (CHI)},
  Xbooktitle    = {Proceedings of the 2019 {CHI} Conference on Human Factors in Computing
                  Systems, {CHI} 2019, Glasgow, Scotland, UK, May 04-09, 2019},
  Xpages        = {3},
  Xpublisher    = {{ACM}},
  year         = {2019},
  Xurl          = {https://doi.org/10.1145/3290605.3300233},
  Xdoi          = {10.1145/3290605.3300233},
  Xtimestamp    = {Sun, 19 Jan 2025 13:11:58 +0100},
  Xbiburl       = {https://dblp.org/rec/conf/chi/AmershiWVFNCSIB19.bib},
  Xbibsource    = {dblp computer science bibliography, https://dblp.org}
}

@article{mehri2026distributional,
  author       = {Shuhaib Mehri and others},
  Xauthor       = {Shuhaib Mehri and
                  Philippe Laban and
                  Sumuk Shashidhar and
                  Marwa Abdulhai and
                  Sergey Levine and
                  Michel Galley and
                  Dilek Hakkani{-}T{\"{u}}r},
  title        = {Measuring and Mitigating the Distributional Gap Between Real and Simulated
                  User Behaviors},
  journal      = {CoRR},
  volume       = {abs/2605.07847},
  year         = {2026},
  Xurl          = {https://doi.org/10.48550/arXiv.2605.07847},
  Xdoi          = {10.48550/ARXIV.2605.07847},
  Xeprinttype   = {arXiv},
  Xeprint       = {2605.07847},
  Xtimestamp    = {Mon, 08 Jun 2026 15:22:43 +0200},
  Xbiburl       = {https://dblp.org/rec/journals/corr/abs-2605-07847.bib},
  Xbibsource    = {dblp computer science bibliography, https://dblp.org}
}

@article{chopra2026personas,
  author       = {Harshita Chopra and
                  Kshitish Ghate and
                  Aylin Caliskan and
                  Tadayoshi Kohno and
                  Chirag Shah and
                  Natasha Jaques},
  title        = {Beyond Cooperative Simulators: Generating Realistic User Personas
                  for Robust Evaluation of {LLM} Agents},
  journal      = {CoRR},
  volume       = {abs/2605.12894},
  year         = {2026},
  Xurl          = {https://doi.org/10.48550/arXiv.2605.12894},
  Xdoi          = {10.48550/ARXIV.2605.12894},
  Xeprinttype   = {arXiv},
  Xeprint       = {2605.12894},
  Xtimestamp    = {Tue, 09 Jun 2026 17:30:38 +0200},
  Xbiburl       = {https://dblp.org/rec/journals/corr/abs-2605-12894.bib},
  Xbibsource    = {dblp computer science bibliography, https://dblp.org}
}

@article{liu2026autoresearchclaw,
  author       = {Jiaqi Liu and others},
  Xauthor       = {Jiaqi Liu and
                  Shi Qiu and
                  Mairui Li and
                  Bingzhou Li and
                  Haonian Ji and
                  Siwei Han and
                  Xinyu Ye and
                  Peng Xia and
                  Zihan Dong and
                  Meng Chen and
                  Congyu Zhang and
                  Letian Zhang and
                  Guiming Chen and
                  Haoqin Tu and
                  Xinyu Yang and
                  Lu Feng and
                  Xujiang Zhao and
                  Haifeng Chen and
                  Jiawei Zhou and
                  Xiao Wang and
                  Weitong Zhang and
                  Hongtu Zhu and
                  Yun Li and
                  Jieru Mei and
                  Hongliang Fei and
                  Jiaheng Zhang and
                  Linjie Li and
                  Linjun Zhang and
                  Yuyin Zhou and
                  Sheng Wang and
                  Caiming Xiong and
                  James Zou and
                  Zeyu Zheng and
                  Cihang Xie and
                  Mingyu Ding and
                  Huaxiu Yao},
  title        = {{AutoResearchClaw}: Self-Reinforcing Autonomous Research with Human-{AI} Collaboration},
  journal      = {CoRR},
  volume       = {abs/2605.20025},
  year         = {2026},
  Xurl          = {https://doi.org/10.48550/arXiv.2605.20025},
  Xdoi          = {10.48550/ARXIV.2605.20025},
  Xeprinttype   = {arXiv},
  Xeprint       = {2605.20025},
  Xtimestamp    = {Fri, 12 Jun 2026 15:09:05 +0200},
  Xbiburl       = {https://dblp.org/rec/journals/corr/abs-2605-20025.bib},
  Xbibsource    = {dblp computer science bibliography, https://dblp.org}
}

@article{elfeki2026hilbench,
  author       = {Mohamed Elfeki and others},
  Xauthor       = {Mohamed Elfeki and
                  Tu Trinh and
                  Kelvin Luu and
                  Guangze Luo and
                  Nathan Hunt and
                  Ernesto Montoya and
                  Nandan Marwaha and
                  Yannis He and
                  Charles Wang and
                  Fernando Crabedo and
                  Alessa Castilo and
                  Bing Liu},
  title        = {{HiL-Bench} (Human-in-Loop Benchmark): Do Agents Know When to Ask for Help?},
  journal      = {CoRR},
  volume       = {abs/2604.09408},
  year         = {2026},
  Xurl          = {https://doi.org/10.48550/arXiv.2604.09408},
  Xdoi          = {10.48550/ARXIV.2604.09408},
  Xeprinttype   = {arXiv},
  Xeprint       = {2604.09408},
  Xtimestamp    = {Fri, 08 May 2026 10:42:59 +0200},
  Xbiburl       = {https://dblp.org/rec/journals/corr/abs-2604-09408.bib},
  Xbibsource    = {dblp computer science bibliography, https://dblp.org}
}

@article{balepur2026dracula,
  author       = {Nishant Balepur and others},
  Xauthor       = {Nishant Balepur and
                  Malachi Hamada and
                  Varsha Kishore and
                  Sergey Feldman and
                  Amanpreet Singh and
                  Pao Siangliulue and
                  Joseph Chee Chang and
                  Rachel Rudinger and
                  Eunsol Choi and
                  Jordan Lee Boyd{-}Graber and
                  Doug Downey and
                  Aakanksha Naik},
  title        = {{DRACULA:} Hunting for the Actions Users Want Deep Research Agents
                  to Execute},
  journal      = {CoRR},
  volume       = {abs/2604.23815},
  year         = {2026},
  Xurl          = {https://doi.org/10.48550/arXiv.2604.23815},
  Xdoi          = {10.48550/ARXIV.2604.23815},
  Xeprinttype   = {arXiv},
  Xeprint       = {2604.23815},
  Xtimestamp    = {Mon, 18 May 2026 08:52:22 +0200},
  Xbiburl       = {https://dblp.org/rec/journals/corr/abs-2604-23815.bib},
  Xbibsource    = {dblp computer science bibliography, https://dblp.org}
}
\bibliographystyle{icml2026/icml2026}

\newpage

\appendix
\section{Evaluator Model Details}
\label{app:evaluator-models}

We use the Collaborative Gym default evaluator (NVIDIA-Nemotron-3-Super-120B) for the tabular-analysis Performance score and the initiative-event classifier used to compute $H_{\mathrm{init,norm}}$, to remain comparable with prior work, and a separate strong labeler (Claude Sonnet~4.6) for the graph-dependent process metrics: Workflow Coverage, Hypothesis Support, and Profile Alignment. We generate reference workflow graphs with Claude Code using Claude Opus~4.6. The generation agent receives the task data, query, benchmark reference hypothesis, and instructions to identify acceptable solution paths, including alternative evidence routes where applicable. We then programmatically check and human-validate the resulting graphs as described in Section~\ref{sec:experimental-setup}.

\section{Simulated-Human Profile Guidance}
\label{app:simulated-human-guidance}

We implement the simulated-human collaborator profiles by adding a static private guidance block $\pi_u$ to the otherwise shared Collaborative Gym simulated-human prompt. This isolates the profile manipulation: both collaborator profiles use the same model, workspace, action loop, communication channel, and own-history visibility described in Section~\ref{sec:experimental-setup}. The profile guidance provides additional background meant to simulate differences in collaborator expertise, work experience, and attention, without assuming perfect task performance. Full code, prompts, profile configuration files, and the reference workflow graphs and validation annotations are available in our code repository at \url{https://github.com/nachiketdk/scaffolded-human-ai-collaboration}.

\paragraph{Data-analysis collaborator.}
The data-analysis profile emphasizes table-grounded evidence work: mapping query concepts to dataset columns, identifying variable families, applying filters or temporal windows, computing extrema or first/second events, validating results, and communicating inspectable evidence.

\paragraph{Researcher collaborator.}
The researcher profile emphasizes domain interpretation and evidential alignment: preserving distinctions such as first versus second, peak versus highest or lowest, increase versus decrease, and bounded-period language; reviewing variable choice, time conversion, ambiguity, and support; noticing missing column mappings or numeric evidence; and shaping final hypotheses around context, variables, result, evidence, and caveats.

\section{Additional Qualitative Trace Figures}
\label{app:additional-qualitative-figures}
These additional trace figures provide the other selected-event pairs referenced in Section~\ref{sec:qualitative-analysis}.

\begin{figure*}[p]
\centering
\setlength{\fboxsep}{4pt}
\begin{subfigure}[t]{0.60\textwidth}
\centering
\abstractTaskSeven
\caption{Task specification}
\end{subfigure}

\vspace{0.7em}
\begin{subfigure}[t]{0.94\textwidth}
\centering
\includegraphics[width=\linewidth]{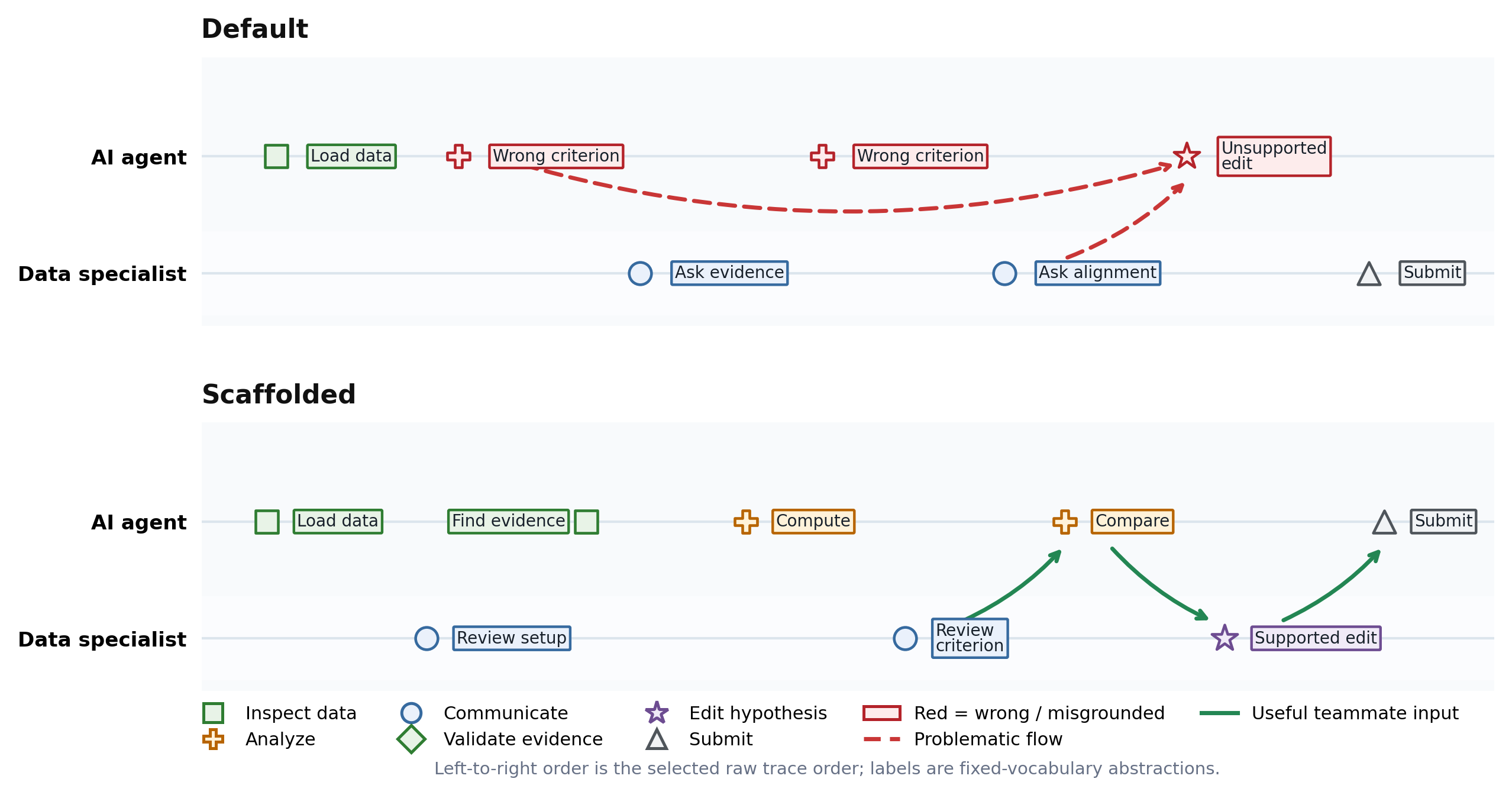}
\caption{Selected-event trace panel}
\end{subfigure}
\caption{Selected trace pair for the copper-peak task, using selected events from the raw trace. The compact tags remove low-level trace text while preserving participant lanes and left-to-right event order. The Default trace carries a wrong criterion into an unsupported hypothesis; the Scaffolded trace shows the data-analysis collaborator's criterion check feeding into comparison and a supported hypothesis.}
\label{fig:appendix-copper-peak-panel}
\end{figure*}

\clearpage

\begin{figure*}[p]
\centering
\setlength{\fboxsep}{4pt}
\begin{subfigure}[t]{0.60\textwidth}
\centering
\abstractTaskThirtyOne
\caption{Task specification}
\end{subfigure}

\vspace{0.7em}
\begin{subfigure}[t]{0.94\textwidth}
\centering
\includegraphics[width=\linewidth]{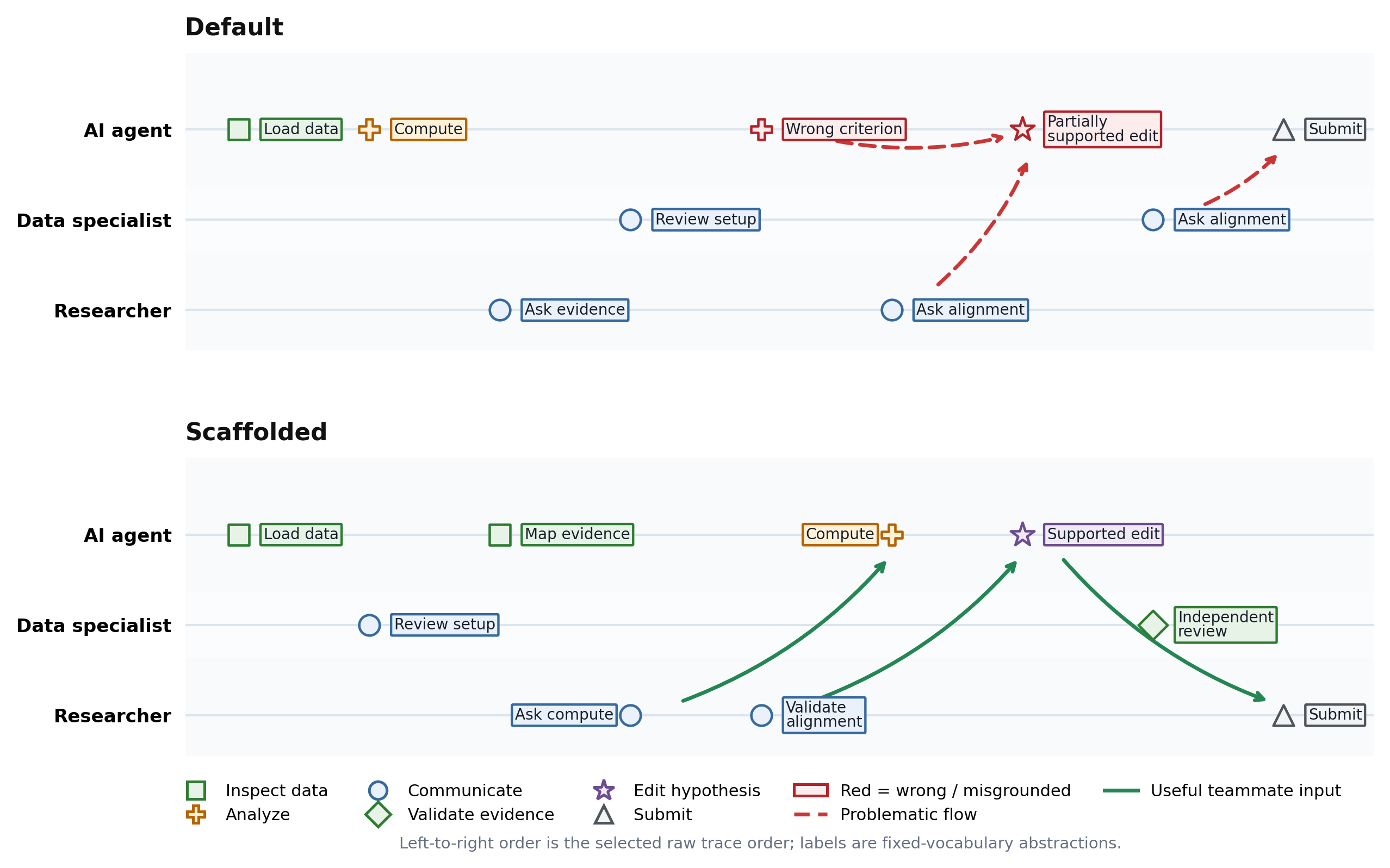}
\caption{Selected-event trace panel}
\end{subfigure}
\caption{Selected trace pair for the pottery-decoration task. The task asks when Diversity in Pottery Decoration begins its highest sustained values. The Default trace anchors the hypothesis to the first elevated plateau rather than the start of the highest sustained plateau. The Scaffolded trace shows teammate checking preceding a more direct computation of the maximum plateau and a more directly supported hypothesis.}
\label{fig:appendix-pottery-decoration-panel}
\end{figure*}

\end{document}